\documentclass{article}

\PassOptionsToPackage{numbers, compress}{natbib}

\usepackage[preprint]{neurips_2024}

\usepackage[utf8]{inputenc}
\usepackage[T1]{fontenc}
\usepackage[colorlinks=true, linkcolor=blue, urlcolor=blue, citecolor=blue, anchorcolor=blue]{hyperref}

\usepackage[dvipsnames, table]{xcolor}
\definecolor{easy_green}{RGB}{147,196,125}
\definecolor{medium_orange}{RGB}{246,178,107}
\definecolor{hard_red}{RGB}{224,102,102}
\definecolor{lightGreen}{RGB}{217,234,211}
\definecolor{lightOrange}{RGB}{252,229,205}
\definecolor{lightRed}{RGB}{244,204,204}

\usepackage{url}
\usepackage{adjustbox}
\usepackage{booktabs}
\usepackage{amsfonts}
\usepackage{nicefrac}
\usepackage{microtype}
\usepackage{bm}
\usepackage{graphicx}
\usepackage{grffile}
\usepackage[hang]{subfigure}
\usepackage{wrapfig}
\usepackage{threeparttable}
\usepackage{bbding}
\usepackage{caption}
\usepackage{amsmath}
\usepackage{multirow}
\usepackage[most]{tcolorbox}
\usepackage{textcomp}
\usepackage{listings}
\usepackage{lipsum}
\usepackage[frozencache,cachedir=.]{minted}

\newcommand{\eg}{\emph{e.g.,}~}
\newcolumntype{L}[1]{>{\raggedright\arraybackslash}p{#1}}

\setcitestyle{square, numbers}

\newcommand{\whitefootnote}[1]{%
  \begingroup
  \renewcommand\thefootnote{\textcolor{white}{\arabic{footnote}}}%
  \hypersetup{hidelinks}
  \footnote{\textcolor{black}{#1}}
  \endgroup
}

\title{\includegraphics[scale=0.05, bb=300 300 700 700]{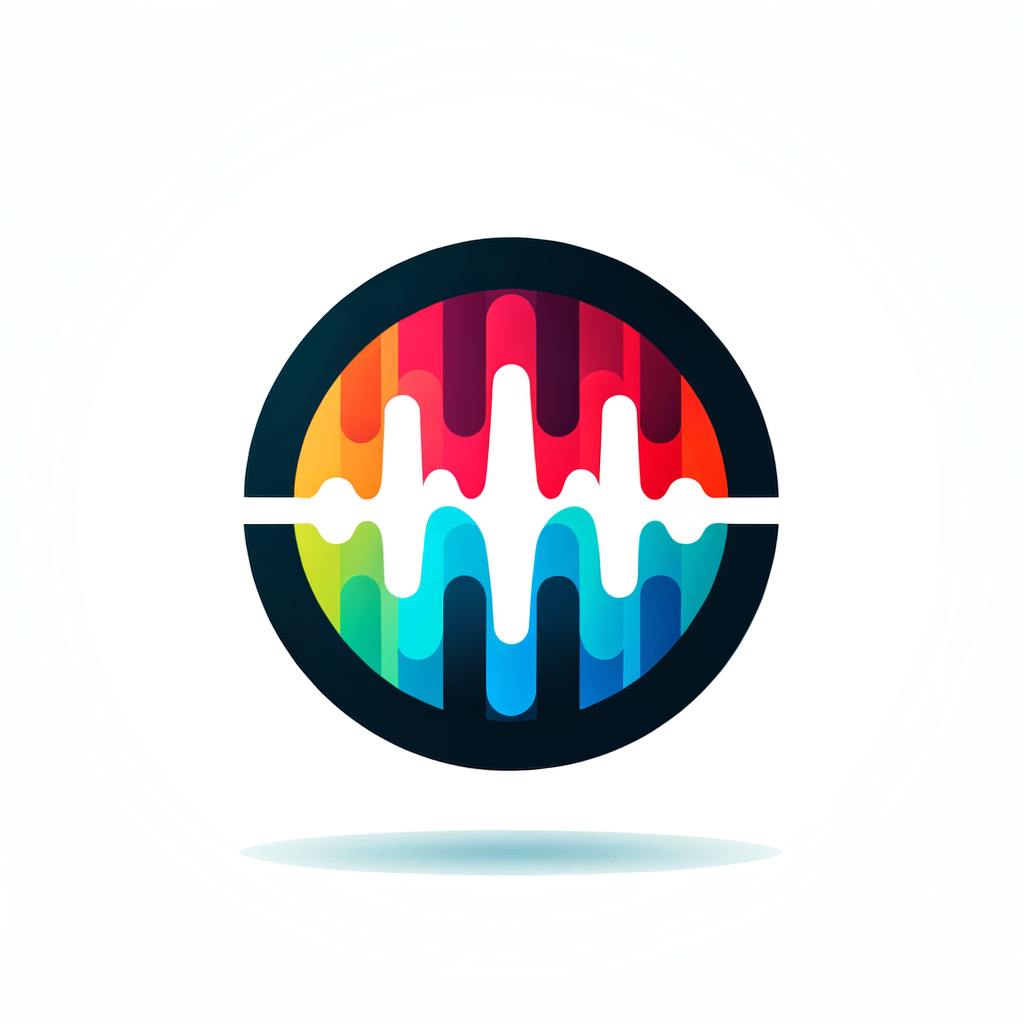}~~AnalogCoder: Analog Circuit Design 
\\  via Training-Free Code Generation}

\author{
   \textbf{Yao Lai}$^1$, \textbf{Sungyoung Lee}$^2$, 
   \textbf{Guojin Chen}$^3$, \textbf{Souradip Poddar}$^2$, \\
   \textbf{Mengkang Hu}$^1$, \textbf{David Z. Pan}$^2$, \textbf{Ping Luo}$^1$ 
\\
$^1$The University of Hong Kong,
$^2$The University of Texas at Austin, \\
$^3$The Chinese University of Hong Kong\\
}

\begin{document}

\whitefootnote{This work is done while the first author is a visiting student at UT Austin.}
\whitefootnote{Corresponding to: Ping Luo (pluo@cs.hku.hk), David Z. Pan (dpan@ece.utexas.edu).}

\maketitle

\begin{abstract}
Analog circuit design is a significant task in modern chip technology, focusing on the selection of component types, connectivity, and parameters to ensure proper circuit functionality. 
Despite advances made by Large Language Models (LLMs) in digital circuit design, the complexity and scarcity of data in analog circuitry pose significant challenges.
To mitigate these issues, we introduce AnalogCoder, the first training-free LLM agent for designing analog circuits through Python code generation.
Firstly, AnalogCoder incorporates a feedback-enhanced flow with tailored domain-specific prompts, enabling the automated and self-correcting design of analog circuits with a high success rate.
Secondly, it proposes a circuit tool library to archive successful designs as reusable modular sub-circuits, simplifying composite circuit creation. 
Thirdly, extensive experiments on a benchmark designed to cover a wide range of analog circuit tasks show that AnalogCoder outperforms other LLM-based methods. 
It has successfully designed 20 circuits, 5 more than standard GPT-4o.
We believe AnalogCoder can significantly improve the labor-intensive chip design process, enabling non-experts to design analog circuits efficiently.
Codes and the benchmark are provided at \href{https://github.com/laiyao1/AnalogCoder}{github.com/laiyao1/AnalogCoder}.
\end{abstract}

\section{Introduction}
Analog circuits, essential for processing real-world signals such as temperature, pressure, sound, and light, are indispensable in modern integrated circuits. 
They facilitate accurate sensing, amplification, and filtering, crucial for linking digital systems with physical environments. 
This functionality underpins reliable data acquisition and signal processing across diverse applications, including wireless communications~\cite{zhang2021wireless}, video sensing~\cite{chatterjee202265nm}, and digital medical devices~\cite{zheng2021self}.

The success of Large Language Models (LLMs)~\cite{zhao2023survey, achiam2023gpt} has brought new opportunities for automatic chip design~\cite{zhong2023llm4eda}. 
Existing related research primarily focuses on two tasks: the generation and correction of Verilog codes~\cite{blocklove2023chip, chang2023chipgpt, thakur2023verigen, thakur2023autochip, fu2023gpt4aigchip, liu2023verilogeval, lu2024rtllm, tsai2023rtlfixer, liu2023rtlcoder, pei2024betterv}, and the writing of design scripts~\cite{wu2024chateda, liu2023chipnemo}.
LLMs can convert natural language descriptions of digital circuit design tasks into Verilog code, a programming language for designing digital circuits. 
Once the code is generated, it can be assessed for correctness by LLMs or human experts, who attempt to fix errors by analyzing error information and simulation outputs~\cite{blocklove2023chip, thakur2023autochip, tsai2023rtlfixer, yao2024hdldebugger}.
Due to the scant representation of Verilog in the public data for training~\cite{guo2024deepseek}, LLMs may not perform as well in generating Verilog code as they do with widely-used programming languages such as C and Python, despite ongoing improvement efforts~\cite{pei2024betterv}.
Similarly, generating design flow scripts is another form of code generation, converting natural language descriptions of design requirements into script files. These scripts, written in Python or Tcl, facilitate the chip design process by invoking APIs at various stages~\cite{ousterhout1993introduction, wu2024chateda, liu2023chipnemo}.
These design flow scripts typically implement straightforward logic to transform fundamental workflows into a series of API calls, akin to control systems used in robotics~\cite{quigley2009ros}.
However, these works are mainly for digital circuit design, as listed in Table~\ref{method_comparison}.
\looseness=-1

\begin{figure}[!t]
    \centering
        \centering
        \includegraphics[width=0.95\textwidth]{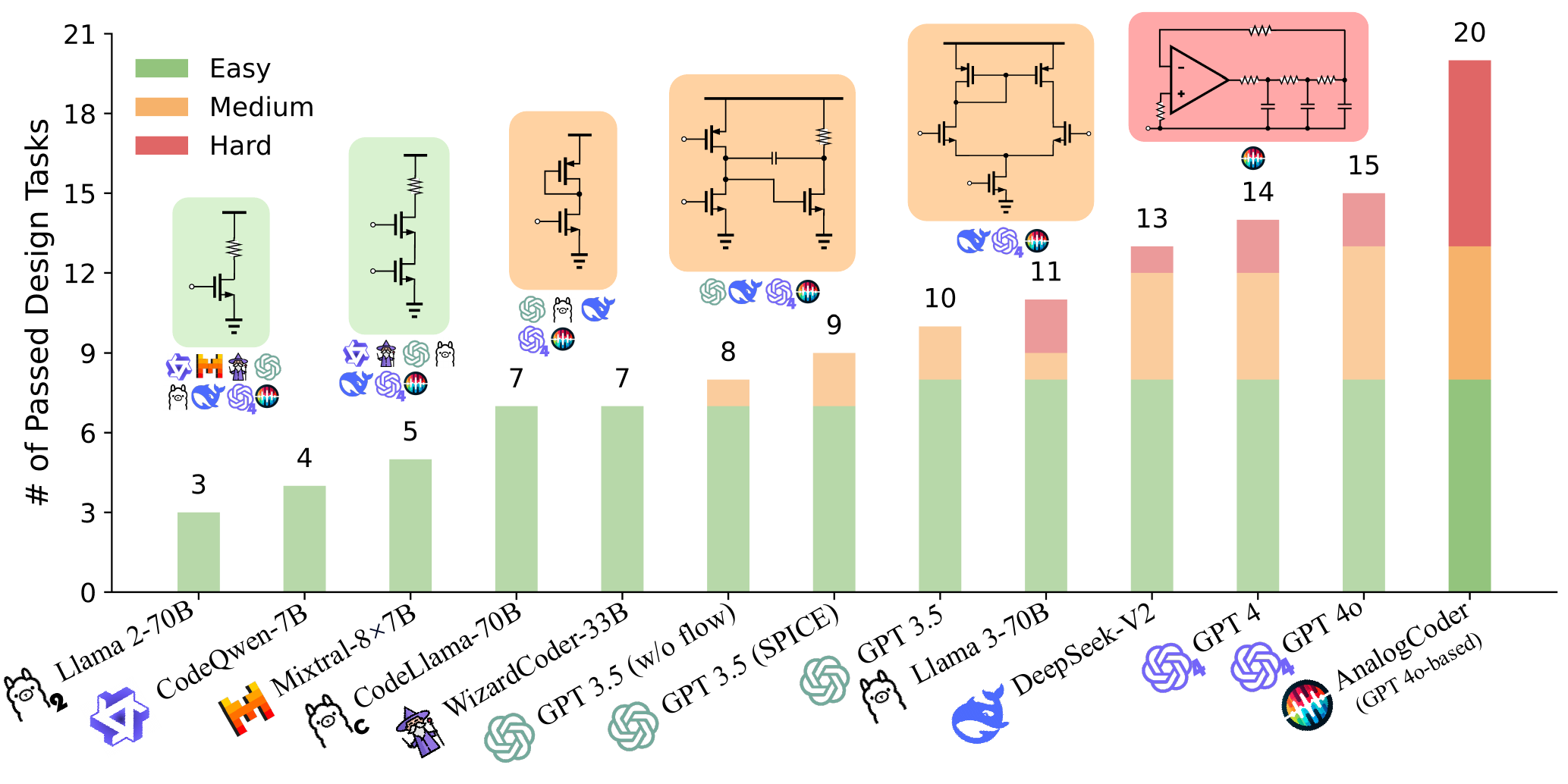} 
        \caption{\small{\textbf{Leaderboard of LLM analog circuit design.} LLMs are ranked by the number of analog circuits they design successfully. Design tasks are classified as easy, medium, or hard based on component count and connection complexity. It displays several designs from our benchmark (Task ID=1, 5, 10, 9, 11, 16) and lists the LLMs that successfully created them. Results are from Table~\ref{main_result} and \ref{ablation}.}}
        \label{leaderboard}
        \vspace{-20pt}
\end{figure}

Analog circuit design presents significantly more challenges than digital circuit design~\cite{johns2008analog, razavi2000design, allen2011cmos}, leaving the field less explored by LLM-aided methods. The primary challenges include:
(1)~\textit{Complexity.} 
Unlike digital circuit design, which predominantly employs simple logic gates, analog circuits comprise diverse components such as voltage and current sources, MOSFETs, resistors, and capacitors.
The complexity is further compounded by the intricate interconnections and settings required. Even minor adjustments can significantly alter the circuit's functionality, potentially leading to a combinatorial explosion due to the vast search space.
(2)~\textit{Abstraction level.} 
Digital circuit design languages like Verilog~\cite{thomas2008verilog} allow developers to write at a high level of abstraction, such as assigning functionality directly, without needing to specify the underlying hardware components like logic gates.
In contrast, analog circuit design requires a direct representation of the physical components in the design code. It necessitates a more detailed and component-specific design process, making it more difficult to utilize LLM assistance effectively.
For example, while a digital adder can be succinctly implemented in a single line of Verilog code, constructing an analog adder requires meticulous configuration and connection of approximately five MOSFETs and three resistors~\cite{chaoui1995cmos}.
(3)~\textit{Corpus data volume.} Although Verilog, used for digital circuit design, constitutes a small fraction (less than 0.1\%) of the repositories on GitHub, SPICE (Simulation Program with Integrated Circuit Emphasis)~\cite{vladimirescu1994spice}, the predominant language for analog design, is even less common. This scarcity suggests that LLMs may find it more challenging to learn the design rules for analog circuits compared to digital ones.
Thus, analog circuit design is a time-intensive, challenging, and error-prone process that predominantly depends on the meticulous contributions of experienced engineers, typically necessitating several days of dedicated expert effort to meet specific functional requirements~\cite{dong2022cktgnn, allen2011cmos}.

To mitigate the shortcomings of traditional manual analog circuit design and to bridge the existing gap in LLM applications for such tasks, we introduce AnalogCoder, a novel training-free LLM-based agent that enables analog circuit design through the generation of Python code.
Specifically, users can describe their desired analog circuit functionalities in natural language, and AnalogCoder automatically generates the corresponding Python code for the designed circuit, leveraging the LLM's strong Python programming capabilities.
To further enhance the design capabilities of LLMs, we propose domain-specific prompt engineering, feedback-enhanced design flow, and the circuit tool library, greatly increasing the success rate of design.

In this work, we prioritize the correct functionality of analog circuits, avoiding extensive parameter optimization, already well-addressed by existing advanced methodologies~\cite{wang2020gcn, lyu2018batch, krylov2023learning}. 
Extensive experiments demonstrate that AnalogCoder can autonomously solve 20 out of 24 analog circuit challenges, as shown in Fig.~\ref{leaderboard}, which surpasses the performance of the standard GPT-4o (15 solved) and the Llama-3 (11 solved).

This paper makes three main \textbf{contributions}:
First, we introduce AnalogCoder, which, to the best of our knowledge, is the \textit{first} LLM-based agent for analog integrated circuit design. 
This agent establishes a new paradigm by generating Python code to design analog circuits.
Second, we develop a feedback-enhanced design flow and a circuit tool library, significantly improving the LLM's ability to design functional analog circuits.
Third, we introduce the \textit{first} benchmark specifically designed to evaluate the ability of LLMs in designing analog circuits. This benchmark comprises 24 unique circuits, three times the number included in the ChipChat benchmark~\cite{chang2023chipgpt} and offers 40\% more circuits than the VeriGen benchmark~\cite{thakur2023verigen}.
It features detailed task descriptions, sample designs, and test-benches, enhancing resources for future research.
\looseness=-1

\begin{table}[!t]
        \caption{\small{\textbf{Comparison of works.} 
AnalogCoder is the first LLM-based work on analog circuit design.
At the same time, AnalogCoder operates without human feedback and features automatic error correction. A comprehensive dataset is developed and provided to evaluate analog circuit design capabilities.
}}
\label{method_comparison}
\resizebox{\textwidth}{!}{
\begin{tabular}{l|cccccc}
\toprule
\textbf{Method} & \textbf{Fully Automated} \textsuperscript{1} & \textbf{Auto Fix Errors} \textsuperscript{2} & \textbf{Benchmark} & \textbf{Open-Source} & \textbf{Training-Free} & \textbf{Circuit Type} \\ 
\midrule
ChipChat \cite{blocklove2023chip}    & ×  & × & \checkmark & \checkmark & \checkmark & Digital \\ 
ChipGPT \cite{chang2023chipgpt}    & ×  & × & \checkmark & × & \checkmark & Digital \\ 
VeriGen  \cite{thakur2023verigen}    & \checkmark & × & \checkmark & \checkmark & ×  & Digital \\
AutoChip  \cite{thakur2023autochip}    & \checkmark & \checkmark & × & \checkmark & \checkmark & Digital \\ 
VerilogEval \cite{liu2023verilogeval} & \checkmark & × & \checkmark & × & × & Digital \\ 
RTLLM \cite{lu2024rtllm} & \checkmark & × & \checkmark & \checkmark & \checkmark & Digital \\ 
RTLfixer  \cite{tsai2023rtlfixer}   & \checkmark & \checkmark & × & \checkmark & \checkmark & Digital \\
RTLCoder  \cite{liu2023rtlcoder}   & \checkmark & × & × & \checkmark & × & Digital \\
ChipNeMo  \cite{liu2023chipnemo}   & \checkmark & × & × & × & × & Digital\textsuperscript{3} \\
BetterV  \cite{pei2024betterv}   & \checkmark & × & × & × & × & Digital \\
\textbf{AnalogCoder} & \checkmark & \checkmark & \checkmark & \checkmark & \checkmark & Analog \\ 
\bottomrule
\end{tabular}
}
{\scriptsize
\textsuperscript{1} Without human involvement. \quad
\textsuperscript{2} Automatic error fix by LLMs. \quad
\textsuperscript{3} Analog circuit only for QA questions.
}
\vspace{-12pt}
\end{table}

\section{Preliminary}

\paragraph{Analog Circuits.}

\begin{figure}[!t]
\centering
\includegraphics[width=\textwidth]{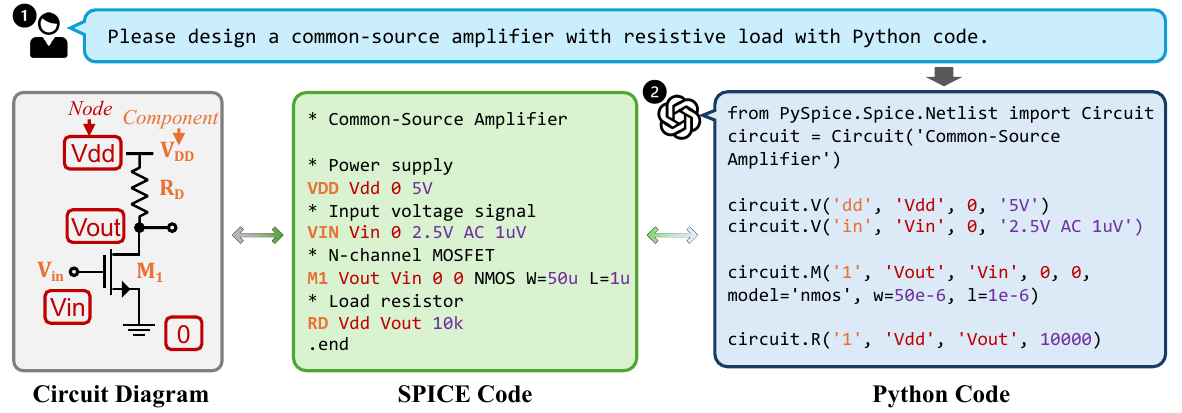}
\caption{\small{\textbf{Representations of designed circuit.} 
We input the design task description into the LLM, which outputs the design in Python. 
Three interconvertible representations of the designed circuit: 
(1) Circuit diagrams, typically take human experts several days to design due to the intricate and demanding process of selecting and connecting components.
(2) SPICE code, which represents circuits through formatted netlists.
(3) Python code with the PySpice library, achieving circuits equivalent to those generated by SPICE code.
However, due to the coding's complexity and non-intuitive nature, experts typically sketch circuit diagrams manually and then generate the codes using design tools.
More code samples are in Appendix Sec.~\ref{example_generated_circuits}.
}}
\label{circuit_example}
\vspace{-15pt}
\end{figure}

Unlike digital circuits, which exclusively process discrete binary signals, analog circuits manage continuous-valued signals, thereby enabling a diverse array of functionalities~\cite{razavidesign}.
For example, an analog amplifier, as depicted in Fig. \ref{circuit_example}, is engineered to enhance the amplitude of an input signal, expressed as \( V_{out}(t) = A_v \times V_{in}(t) \), where \( V_{in}(t) \) and \( V_{out}(t) \) denote the time-variant behavior of the input and output voltage signals, respectively, and \( A_v \) represents the voltage gain of the amplifier. 
Moreover, operational amplifiers (op-amps) are high-gain voltage amplifiers with differential inputs, featuring a non-inverting input \( V_{inp} \) and an inverting input \( V_{inn} \). 
The output of the op-map is expressed as \( V_{out}(t) = A_v \times [V_{inp}(t) - V_{inn}(t)] \), enabling it to be configured to perform a variety of analog signal operations, including integration, differentiation, addition, and subtraction. 
When configured as an adder, for instance, the operational amplifier can implement the function \( V_{out}(t) = -[V_{in1}(t) + V_{in2}(t)] \).
Moreover, when set up as an integrator, it can integrate the input voltage signal, yielding an output given by \( V_{out}(t) = - \int V_{in}(t) \, dt / \tau\), where $\tau$ represents the time constant associated with the resistance and capacitance values within the circuit.
These analog circuits demonstrate how analog operations transform input signals into output signals, accomplishing computations more efficiently than clock-dependent digital circuits.
Testing an analog circuit involves applying a specific input and verifying that the output aligns with expected standards to ensure its correct operation. 
After simulation, attributes such as gain, common-mode gain, and phase difference are identified as specifications. 
These specifications are essential for validating the circuit's performance against its intended design requirements.

\paragraph{Code Representation for Circuits.} 
To facilitate the description and simulation of analog circuit designs, SPICE~\cite{vladimirescu1994spice} has been introduced. 
Often used as a programming language, this tool allows designers to specify the complex interconnections between electronic components within a circuit.
With SPICE codes, the behavior of circuits can be accurately simulated and analyzed, with each component, such as resistors, capacitors, and voltage or current sources, carefully itemized and connected in a notation recognized industry-wide.
In the SPICE syntax, the fundamental constructs are elements and nodes (see Fig.~\ref{circuit_example}).
The elements refer to various electronic components like resistors and transistors, while nodes denote the points at which these elements are interconnected.
As shown in the circuit diagram in Fig. \ref{circuit_example}, the amplifier comprises four elements: two voltage sources, $V_{dd}$ and $V_{in}$, for the power supply and signal input, respectively; one N-channel MOSFET, $M_1$, for signal amplification; and one resistive load, $R_D$. 
Four lines in the SPICE code describe these elements.
Each line in the SPICE code starts with the element name, followed by the names of the nodes to which the element is connected.
For instance, the resistor $R_D$ is connected between nodes $V_{dd}$ and $V_{out}$. The corresponding SPICE code line is `\texttt{RD Vdd Vout 10k}', where `\texttt{10k}' denotes $10\,\text{k}\Omega$. 
Specifically, since a MOSFET has four connection nodes, the corresponding code line will include four node labels delineating the drain, gate, source, and bulk connections.
PySpice~\cite{salvaire2021pyspice} integrates SPICE code with the Python programming language, leveraging Python's user-friendly syntax and robust ecosystem to simplify circuit simulation and data processing, as demonstrated in the Python code in Fig.~\ref{circuit_example}.
This integration allows for more accessible and efficient design workflows, broadening the usability of SPICE.
Since LLMs excel at Python programming~\cite{khan2023xcodeeval, zheng2023codegeex}, we chose Python with the PySpice library to automate the creation of circuits, replacing the tedious manual process.

\section{Our Approach}

\begin{figure}[!t]
\centering
\includegraphics[width=\textwidth]{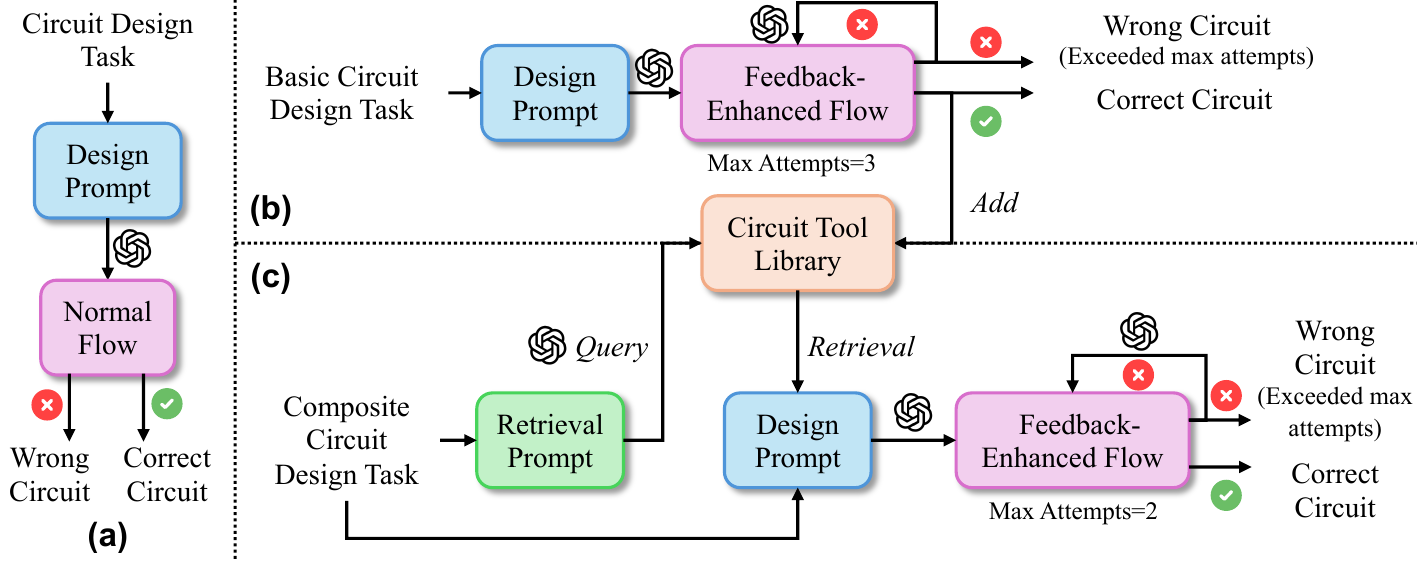}
\caption{\small{\textbf{Method Overview.} \textbf{(a)
Standard method.}
By converting circuit design tasks into prompts and inputting the LLM's outputs into the normal flow, the correctness of the circuit is verified.
\textbf{(b) Our method for basic circuit design.}
Input the design prompts to the feedback-enhanced design flow, enabling the automated error fix with LLMs.
Successfully designed circuits are added to the circuit tool library, while failed designs are returned to the LLM for automatic fixing.
 \textbf{(c) Our method for composite circuit design.} The process adds a step of querying the library to retrieve invocation methods for subcircuits, which are then integrated into the design prompt to facilitate the design of composite circuits.
}}
\label{method_overview}
\vspace{-15pt}
\end{figure}

\paragraph{Method Overview.} 
AnalogCoder is an LLM-based agent that interprets task descriptions in natural language to automatically generate Python code, representing functionally correct analog circuits.
To enhance the design capabilities of the agent, we implemented a comprehensive methodology as shown in Fig.~\ref{method_overview},  including prompt engineering, a feedback-enhanced design flow, and a circuit tool library. 
Prompt engineering enhances the agent's design thinking through strategic, problem-solving prompts.
The feedback-enhanced design flow uses multiple checks to provide error feedback to the agent, facilitating the correction of failed designs by LLMs. 
The circuit tool library, a modular sub-circuit repository, systematically organizes designed circuits as tools, enabling straightforward retrieval and reuse by LLMs for complex circuit designs.

\paragraph{Prompt Engineering.} 
We initially established a well-crafted design prompt to maximize the design capabilities of LLMs. 
Our approach to prompt engineering encompasses three main aspects: 
(1) \textit{programming language selection}, (2) \textit{in-context learning}~\cite{dong2022survey}, and (3) \textit{Chain-of-Thought}~\cite{wei2022chain}.
Despite the capability of LLMs to generate code in multiple programming languages, their performance in Python surpasses that in most others~\cite{cassano2023multipl, zheng2023codegeex}.
Additionally, many prominent code-generating LLMs, such as CodeLlama~\cite{roziere2023code} and WizardCoder~\cite{luo2023wizardcoder}, are primarily fine-tuned on Python datasets, indicating a bias towards Python.
Conversely, the training datasets for most LLMs, which are based on GitHub, do not contain sufficient data on SPICE code~\cite{guo2024deepseek}. 
Therefore, to mitigate this limitation, we directly prompt the LLM to generate executable Python code compatible with the PySpice library.
Furthermore, we integrate in-context learning~\cite{dong2022survey} to enhance circuit design, providing a detailed example of a two-stage amplifier with active and resistor loads as one-shot learning~\cite{brown2020language}.
This example facilitates the LLM's learning and imitation and standardizes its output, minimizing errors.
All design tasks are distinct from the provided example to maintain evaluation fairness.
Additionally, the Chain-of-Thought strategy~\cite{wei2022chain} involves prompting the LLM to generate a detailed design plan, including necessary components and their interconnections.
This plan subsequently guides the generation of the corresponding design code, simplifying the design task significantly.
The complete prompt template can be seen in Appendix Sec.~\ref{complete_prompt}.

\begin{figure}[!t]
  \centering
  \includegraphics[width=\textwidth]{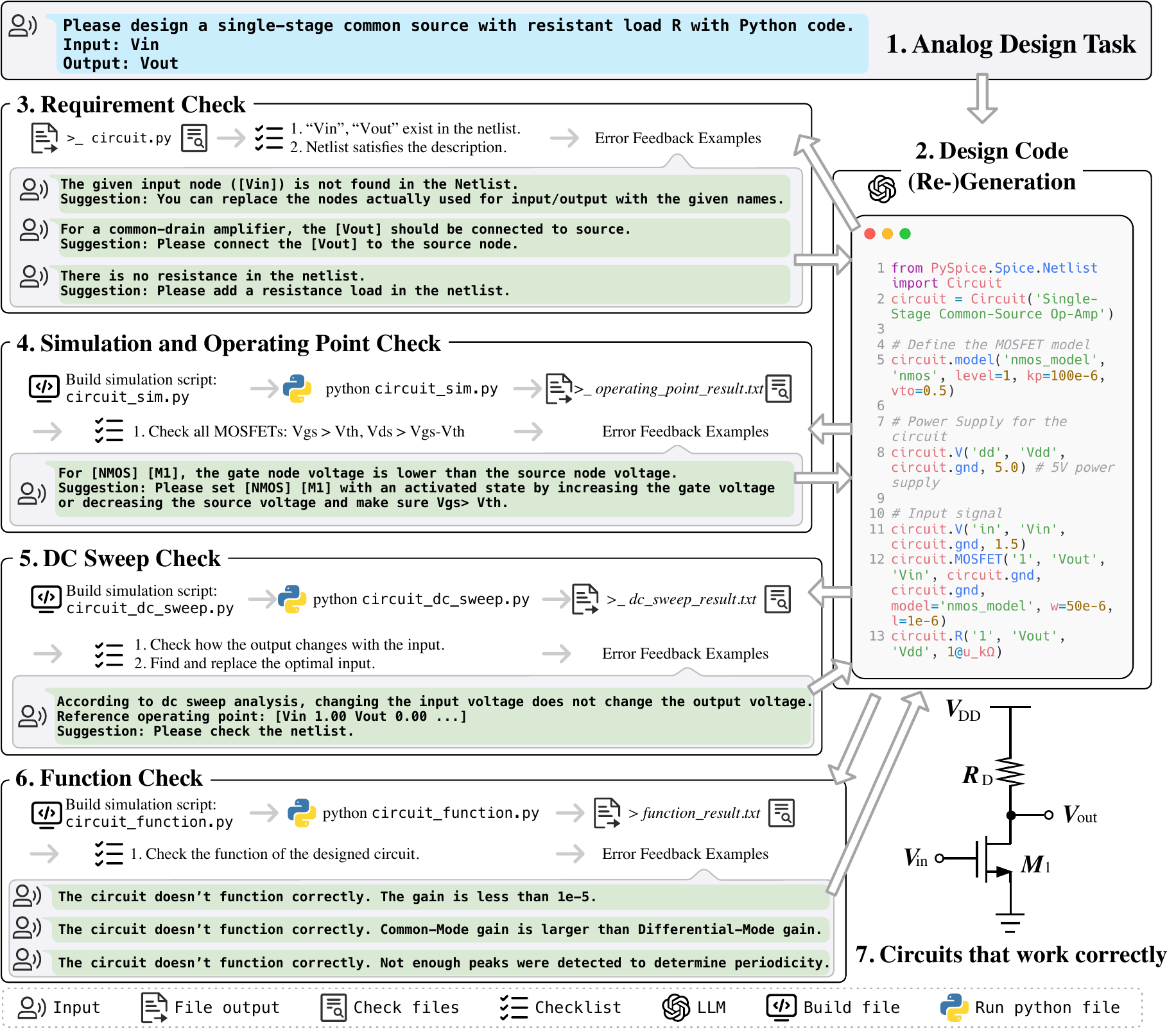}
  \caption{\small{\textbf{Feedback-Enhanced Design Flow.} The flow facilitates autonomous error correction in designs by the LLM agent without human intervention. Error messages identified during the checking process are returned to the LLM to assist in refining the design. The entire flow is adaptable to nearly all categories of analog circuits.}}
  \label{design_flow}
  \vspace{-10pt}
\end{figure}

\paragraph{Feedback-Enhanced Design Flow.}
Various errors are often observed in the code generated by LLMs.
Consequently, guiding LLMs to correct the generated codes based on error messages is crucial. 
Numerous studies \cite{hong2023metagpt, chen2023teaching, olausson2023self, tsai2023rtlfixer, thakur2023verigen} have suggested that providing LLMs with relevant error information helps LLMs fix faulty code.
However, for the analog circuit design, besides the runtime errors that may occur when executing SPICE simulations, additional verification of circuit-related information is necessary to ensure the correctness of the design.
In analog circuit design, when a design fails, we return either runtime errors from the Python code or circuit-specific test errors to the LLM, as illustrated in Fig.~\ref{design_flow}.
We divide the feedback-enhanced flow into four stages: 
(1) \textit{requirement check}, (2) \textit{simulation and operating point check}, (3) \textit{DC sweep check}, and (4) \textit{function check}.
The \textit{requirement check} is to verify whether the generated code meets the basic design requirements, such as the presence of requisite inputs and outputs, and the inclusion of essential circuit components.
The \textit{simulation and operating point check} initially assesses whether the generated analog circuit can successfully execute simulations, aiming to identify issues such as floating nodes and other potential errors.
Once the simulation passes, the static operating point voltages of nodes are achieved. 
Examining these operating point voltages ensures that the MOSFET transistors are in their correct operational states.
The \textit{DC sweep check} performs a direct current (DC) analysis by changing the voltage at the input nodes and observing the corresponding changes at the output nodes to verify the integrity of the signal path from input to output.
This method also helps identify the optimal bias voltage, increasing the success rate of the design.
The \textit{function check} simulates specific input waveforms and observes the outputs to verify the analog circuit's fundamental functionalities as Appendix Table~\ref{criteria}. 
The simulation may involve DC, AC (alternating current), or transient analyses depending on the circuit types.
For any errors occurring in these checks, the relevant error information is returned to the LLM, which then regenerates a circuit design. 
Due to limitations in the LLMs' code repair capabilities, we allow up to three code generations, which means up to two retries.

\begin{figure}[!t]
  \centering
  \includegraphics[width=\textwidth]{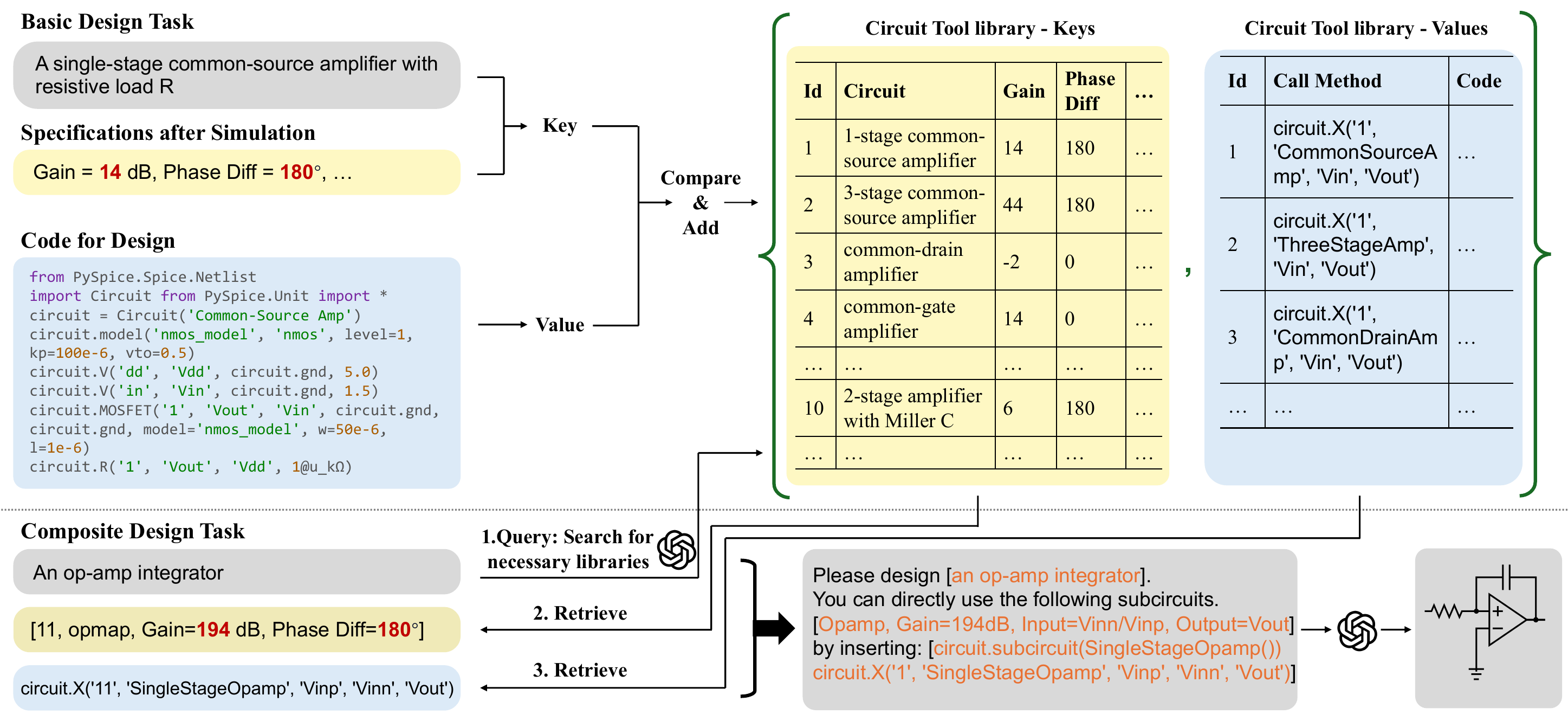}
  \caption{\small{\textbf{Circuit Tool library.} \textbf{Top:} Addition of new tools derived from successfully designed basic circuits. Here, descriptions and specifications are keys, while design codes are stored as values.
\textbf{Bottom:} Retrieval of tools from the library for designing composite circuits. The process begins with the LLM querying necessary tools using the task description. Subsequently, the keys and values of the retrieved tools, with the task description, are employed as prompts for circuit design.
}}
  \label{skill_library}
  \vspace{-10pt}
\end{figure}

\paragraph{Circuit Tool Library.} 
As analog circuit design tasks become more complex and the implementation code grows more intricate, it becomes increasingly challenging for LLMs to generate correct circuits. 
To address this complexity, basic circuits can be encapsulated into subcircuit modules in the SPICE code, facilitating their integration into more composite assemblies. 
Building on this modular approach and inspired by the tool-based LLM studies~\cite{wang2023voyager, qin2023toolllm}, we adopted a circuit tool library that stores correctly designed subcircuits for easy reuse in more complex designs.
As illustrated in Fig.~\ref{skill_library}, our approach involves two main processes: adding circuits to the library (top) and retrieving circuits from the library (bottom). After an LLM-based agent completes a basic circuit design task, we add the circuit codes and the specifications from the simulation results to the circuit tool library. 
If a circuit task has been successfully completed multiple times, store the optimal circuit design based on the key specification, such as gain.
The task descriptions and circuit information are stored as keys for queries, while the codes and calling methods are stored as values.
In composite circuit design, the task description is used to formulate a query prompt, enabling the retrieval of the requisite subcircuit tools by LLMs. 
The agent initially retrieves the indices of the required subcircuits and then uses these indices to fetch all corresponding specifications and calling methods. 
This information is then integrated with the task description and automatically re-entered into the LLM to design the circuit. 
At this stage, the agent uses the retrieved subcircuits' calling methods to directly integrate them into the code, thereby designing composite circuits.
As shown in Fig. \ref{skill_library}, when designing an op-amp integrator, the LLM queries and retrieves the index corresponding to the required subcircuit, a single-stage op-amp. 
Subsequently, the task description, along with the pertinent information of this subcircuit, is input into the LLM, which then generates the design code for the op-amp integrator.

\paragraph{Fine-tuning.} Due to the scarcity of datasets for analog circuits and inspired by GPT-assisted data generation~\cite{liu2024visual}, we collected samples of successful circuit designs created by GPT-3.5, GPT-4o, and Llama-3 to fine-tune GPT-3.5 by the provided API. 
We gathered successful designs for each task and clustered them into three categories using text vectorization TF-IDF~\cite{sparck1972statistical}. 
One design from each category was selected and paired with the input prompt to form initial pairs, then refined through text filtering to create the fine-tuning data. Cross-validation techniques~\cite{kohavi1995study} are applied to ensure that tasks used in fine-tuning were excluded from the testing set. Further details are provided in Appendix Sec.~\ref{experimental_settings}.
\looseness=-1

\section{Experiments}

We extensively evaluate the capability of LLMs in analog circuit design, including Mixtral-7×8B~\cite{jiang2024mixtral}, CodeLlama-70B-Instruct~\cite{roziere2023code}, Wizardcoder-33B-V1.1~\cite{luo2023wizardcoder}, Llama3-70B~\cite{llama3modelcard}, DeepSeek-V2~\cite{deepseekai2024deepseekv2}, GPT-3.5-turbo~\cite{brown2020language}, GPT-4-turbo~\cite{achiam2023gpt} and GPT-4o. 
CodeLlama and WizardCoder are code generation LLMs, fine-tuned on Llama2~\cite{touvron2023llama} and StarCoder~\cite{li2023starcoder}, respectively. 
Llama-3 and DeepSeek-V2 are the newest open-source general LLMs.
WizardCoder, DeepSeek-V2, and Llama-3 are LLMs that outperformed GPT-3.5 on the HumanEval~\cite{chen2021evaluating} coding tasks~\cite{liu2024your}.
For additional models, see Appendix Sec.~\ref{supplementray_exp}.
Open-source models were evaluated on 4 Nvidia A100 GPUs.

\paragraph{Metrics.}
We use `Pass@k'~\cite{kulal2019spoc} (k=1, 5), a metric widely used in code generation tasks~\cite{roziere2023code, li2022competition, luo2023wizardcoder, liu2024your, guo2024deepseek}, as the main evaluation metric. 
It is defined as the ratio of correct generations within $k$ independent trials, with higher values being better.
We conduct $n$ trials $(n\geq k)$, and compute $\textit{Pass@k}=1-\tbinom{n-c}{k}/\tbinom{n}{k}$, where $c$ is the number of successful trials.
For open-source LLMs and GPT-3.5, we set $n=30$; for fine-tuned GPT-3.5, GPT-4, and  GPT-4o, $n=15$.
`Number of solved' refers to the count of distinct tasks for which a circuit design is successfully achieved at least once in $n$ trials.
\looseness=-1

\paragraph{Benchmark.} 
We have developed a comprehensive benchmark of analog circuit design tasks, detailed in Table~\ref{benchmark}, to fill the gap in open-source benchmarks for this field. The difficulty of these tasks is determined by the number of components and the complexity of their connections. Tasks 1-15 are basic circuits, while 16-24 are composite circuits. Details are available in Appendix Sec.~\ref{sec:benchmark_detail}.
\looseness=-1

\begin{table}[htbp]
\caption{\small{\textbf{Benchmark Descriptions.} All analog circuit design tasks are listed with their corresponding types. Different difficulties are distinguished by background colors (\textcolor{easy_green}{\textbf{easy}}, \textcolor{medium_orange}{\textbf{medium}}, and \textcolor{hard_red}{\textbf{hard}}).}}
\label{benchmark}
\begin{adjustbox}{max width=\textwidth}
\begin{tabular}{r|l|l|r|l|l}
\toprule
\textbf{Id} & \textbf{Type} & \textbf{Circuit Description} & \textbf{Id} & \textbf{Type} & \textbf{Circuit Description} \\
\midrule
\rowcolor{lightGreen} 
1  & Amplifier & Common-source amp. with R load & \cellcolor{lightOrange}13 & \cellcolor{lightOrange}Opamp & \cellcolor{lightOrange}Common-source op-amp with R loads \\
\rowcolor{lightGreen} 
2  & Amplifier & 3-stage common-source amplifier with R loads & \cellcolor{lightRed}\cellcolor{lightRed}14 & \cellcolor{lightRed}Opamp & \cellcolor{lightRed}2-stage op-amp with active loads  \\
\rowcolor{lightGreen} 
3  & Amplifier & Common-drain amp. with R load & \cellcolor{lightRed}15 & \cellcolor{lightRed}Opamp & \cellcolor{lightRed}Cascode op-amp with cascode loads\\
\rowcolor{lightGreen} 
4  & Amplifier & Common-gate amp. with R load & \cellcolor{lightRed}16 & \cellcolor{lightRed}Oscillator & \cellcolor{lightRed}Wien Bridge oscillator \\
\rowcolor{lightGreen} 
5  & Amplifier & Cascode amp. with R load & \cellcolor{lightRed}17 & \cellcolor{lightRed}Oscillator & \cellcolor{lightRed}RC Shift oscillator \\
\rowcolor{lightGreen} 
6  & Inverter  & NMOS inverter with R load & \cellcolor{lightRed}18 & \cellcolor{lightRed}Integrator & \cellcolor{lightRed}Op-amp integrator \\
\rowcolor{lightGreen} 
7  & Inverter  & Logical inverter with NMOS and PMOS & \cellcolor{lightRed}19 & \cellcolor{lightRed}Differentiator & \cellcolor{lightRed}Op-amp differentiator \\
\rowcolor{lightGreen} 
8  & Current Mirror & NMOS constant current source with R load & \cellcolor{lightRed}20 & \cellcolor{lightRed}Adder & \cellcolor{lightRed}Op-amp adder \\
\rowcolor{lightOrange} 
9  & Amplifier & Common-source amp. with diode-connected load & \cellcolor{lightRed}21 & \cellcolor{lightRed}Subtractor & \cellcolor{lightRed}Op-amp subtractor \\
\rowcolor{lightOrange} 
10 & Amplifier & 2-stage amplifier with Miller compensation C & \cellcolor{lightRed}22 & \cellcolor{lightRed}Schmitt trigger & \cellcolor{lightRed}Non-inverting Schmitt trigger \\
\rowcolor{lightOrange} 
11 & Opamp & Op-amp with active current mirror loads & \cellcolor{lightRed}23 & \cellcolor{lightRed}VCO & \cellcolor{lightRed}Voltage-Controlled Oscillator \\
\rowcolor{lightOrange}  
12 & Current Mirror & Cascode current mirror & \cellcolor{lightRed}24 & \cellcolor{lightRed}PLL & \cellcolor{lightRed}Phase-Locked Loop \\
\bottomrule
\end{tabular}
\end{adjustbox}
\vspace{-5pt}
\end{table}

\paragraph{Main Results.} 
Table~\ref{main_result} compares our LLM agent, AnalogCoder, which is based on GPT-4o and incorporates prompt engineering, flow feedback, and a circuit tool library with other LLM-based methods.
To ensure a fair comparison and highlight the tool library's impact, we applied our strategies across all LLMs but specifically excluded the circuit tool library from GPT-4o to isolate its effects.
The results indicate that Llama-3 and DeepSeek-V2, the latest open-source models, demonstrate a marginally superior capability in circuit design compared to GPT-3.5. 
However, other open-source models still exhibit a certain gap compared to GPT-3.5, although some surpassed GPT-3.5 in normal Python coding tasks \cite{liu2024your}.
This is primarily because circuit design requires both coding skills and specific background knowledge; hence, general LLMs tend to perform better.
GPT-4o is still the best LLM for analog circuit design, generally consistent with other findings on its performance in coding tasks \cite{luo2023wizardcoder, bai2023qwen, liu2024your, guo2024deepseek}. 
Benefiting from the circuit tool library, GPT-4o and Llama-3 can further utilize existing circuits to design more challenging composite circuits, significantly enhancing their design capabilities.
More results can be seen in Appendix Sec.~\ref{supplementray_exp}.

\begin{table}[!t]
	\caption{\small{\textbf{Main results.} All LLMs, except GPT-4o, have been enhanced by prompt engineering, design flow feedback, and the circuit tool library. To highlight the impact of the circuit tool library, GPT-4o was evaluated without it. AnalogCoder can be seen as the GPT-4o enhanced with the circuit tool library.}}
	\label{main_result}
    \centering
    \begin{adjustbox}{max width=\textwidth}
    \begin{tabular}{r|rr|rr|rr|rr|rr|rr|rr}
\toprule
\textbf{Model} & \multicolumn{2}{c|}{\textbf{CodeLlama-70B}} & \multicolumn{2}{c|}{\textbf{WizardCoder-33B}} & \multicolumn{2}{c|}{\textbf{DeepSeek-V2}} & \multicolumn{2}{c|}{\textbf{Llama3-70B}} & \multicolumn{2}{c|}{\textbf{GPT3.5}} & \multicolumn{2}{c|}{\textbf{GPT4o (w/o tool)}} & \multicolumn{2}{c}{\textbf{AnalogCoder}} \\
Task ID & Pass@1 & Pass@5 & Pass@1 & Pass@5 & Pass@1 & Pass@5 & Pass@1 & Pass@5 & Pass@1 & Pass@5 & Pass@1 & Pass@5 & Pass@1 & Pass@5 \\
\midrule
1  & 20.0 & 70.2 & 93.3 & \textbf{100.0} & \textbf{100.0} & \textbf{100.0} & 93.3 & \textbf{100.0} & 86.7 & \textbf{100.0} & \textbf{100.0} & \textbf{100.0} & \textbf{100.0} & \textbf{100.0} \\
2  &  3.3 & 16.7 & 13.3 & 53.8 & 93.3 & \textbf{100.0} & 20.0 & 70.2 & 70.0 & 99.9 & \textbf{100.0} & \textbf{100.0} & \textbf{100.0} & \textbf{100.0} \\
3  &  0.0 &  0.0 &  0.0 &  0.0 & 83.3 & \textbf{100.0} & 90.0 & \textbf{100.0} &  3.3 & 16.7 & \textbf{100.0} & \textbf{100.0} & \textbf{100.0} & \textbf{100.0} \\
4  &  3.3 & 16.7 & 10.0 & 43.3 & 70.0 & 99.9 & 83.3 & \textbf{100.0} & 50.0 & 97.9 & \textbf{100.0} & \textbf{100.0} & \textbf{100.0} & \textbf{100.0} \\
5  &  3.3 & 16.7 & 13.3 & 53.8 & 76.7 & \textbf{100.0} & 20.0 & 70.2 & 10.0 & 43.3 & \textbf{100.0} & \textbf{100.0} & \textbf{100.0} & \textbf{100.0} \\
6  & 23.3 & 76.4 & 13.3 & 53.8 & \textbf{100.0} & \textbf{100.0} & \textbf{100.0} & \textbf{100.0} & 73.3 & \textbf{100.0} & \textbf{100.0} & \textbf{100.0} & \textbf{100.0} & \textbf{100.0} \\
7  & 10.0 & 43.3 &  6.7 & 31.0 & \textbf{100.0} & \textbf{100.0} & \textbf{100.0} & \textbf{100.0} & 76.7 & \textbf{100.0} & \textbf{100.0} & \textbf{100.0} & \textbf{100.0} & \textbf{100.0} \\
8  & 13.3 & 53.8 & 20.0 & 70.2 & 96.7 & \textbf{100.0} & 93.3 & \textbf{100.0} & 66.7 & 99.8 & \textbf{100.0} & \textbf{100.0} & \textbf{100.0} & \textbf{100.0} \\
9  &  0.0 &  0.0 &  0.0 &  0.0 & 93.3 & \textbf{100.0} &  0.0 &  0.0 & 30.0 & 85.7 & \textbf{100.0} & \textbf{100.0} &\textbf{100.0} & \textbf{100.0} \\
10 &  0.0 &  0.0 &  0.0 &  0.0 & \textbf{100.0} & \textbf{100.0} &  83.3 & \textbf{100.0} & 46.7 & 96.9 & \textbf{100.0} & \textbf{100.0} & \textbf{100.0} & \textbf{100.0} \\
11 & 0.0 & 0.0 & 0.0 & 0.0 & 3.3 & 16.7 & 0.0 & 0.0 & 0.0 & 0.0 & \textbf{100.0} & \textbf{100.0} & \textbf{100.0} & \textbf{100.0} \\
12 & 0.0 & 0.0 & 0.0 & 0.0 & 0.0 & 0.0 & 0.0 & 0.0 & 0.0 & 0.0 & \textbf{13.3} & \textbf{57.1} & \textbf{13.3} & \textbf{57.1}  \\
13 & 0.0 & 0.0 & 0.0 & 0.0 & 3.3 & 16.7 & 0.0 & 0.0 & 0.0 & 0.0 & \textbf{100.0} & \textbf{100.0} & \textbf{100.0} & \textbf{100.0} \\
14 & 0.0 & 0.0 & 0.0 & 0.0 & 6.7 & 31.0 & 0.0 & 0.0 & 0.0 & 0.0 & \textbf{73.3} & \textbf{100.0} & \textbf{73.3} & \textbf{100.0} \\
15 & 0.0 & 0.0 & 0.0 & 0.0 & 0.0 & 0.0 & 0.0 & 0.0 & 0.0 & 0.0 & \textbf{13.3} & \textbf{57.1} & \textbf{13.3} & \textbf{57.1} \\
16 & 0.0 & 0.0 & 0.0 & 0.0 & 0.0 & 0.0 & 0.0 & 0.0 & 0.0 & 0.0 & 0.0 & 0.0 & \textbf{6.7} & \textbf{33.3} \\
17 & 0.0 & 0.0 & 0.0 & 0.0 & 0.0 & 0.0 & 0.0 & 0.0 & 0.0 & 0.0 & 0.0 & 0.0 & 0.0 & 0.0 \\
18 & 0.0 & 0.0 & 0.0 & 0.0 & 0.0 & 0.0 & 3.3 & 16.7 & 0.0 & 0.0 & 0.0 & 0.0 & \textbf{100.0} & \textbf{100.0} \\
19 & 0.0 & 0.0 & 0.0 & 0.0 & 0.0 & 0.0 & 0.0 & 0.0 & 0.0 & 0.0 & 0.0 & 0.0 & \textbf{60.0} & \textbf{99.8} \\
20 & 0.0 & 0.0 & 0.0 & 0.0 & 0.0 & 0.0 & 3.3 & 16.7 & 0.0 & 0.0 & 0.0 & 0.0 & \textbf{100.0} & \textbf{100.0} \\
21 & 0.0 & 0.0 & 0.0 & 0.0 & 0.0 & 0.0 & 0.0 & 0.0 & 0.0 & 0.0 & 0.0 & 0.0 & \textbf{20.0} & \textbf{73.6} \\
22-24 & 0.0 & 0.0 & 0.0 & 0.0 & 0.0 & 0.0 & 0.0 & 0.0 & 0.0 & 0.0 & 0.0 & 0.0 & 0.0 & 0.0 \\
\midrule
Avg & 3.2 & 12.2 & 7.1 & 16.9 & 38.6 & 44.3 & 28.8 & 36.4 & 21.4 & 35.0 & 54.2 & 58.9 & \textbf{66.1} & \textbf{75.9} \\
\# Solved & 7 & 7 & 7 & 7 & 13 & 13 & 11 & 11 & 10 & 10 & 15 & 15 & \textbf{20} & \textbf{20} \\
\bottomrule
\end{tabular}
    \end{adjustbox}
\vspace{-10pt}
\end{table}

\paragraph{Ablations.} 
We evaluated the effectiveness of various components within our approach using the GPT-3.5 model, with results presented in Table \ref{ablation}.
Specifically, ``GPT-3.5 (SPICE)'' refers to the GPT-3.5 in which the LLM is prompted to generate SPICE codes rather than Python. 
The variants ``GPT-3.5 (w/o context)'' and ``GPT-3.5 (w/o CoT)'' explore the impact on performance when omitting in-context information and Chain-of-Thought reasoning from the prompts, respectively.
Furthermore, ``GPT-3.5 (w/o flow)'' indicates a setup in which our proposed design flow was not utilized, and only the first generated codes were applied for functional testing. 
The findings consistently show that removing these components leads to a decrease in design performance.

\begin{table}[!t]
	\caption{\small{\textbf{Ablation study and fine-tuning.} A series of ablation studies on the GPT-3.5 model validate the efficacy of the proposed method by systematically removing components of our approach.
    Fine-tuned GPT-3.5 improves the success rate of designs but does not increase the number of successful circuit designs.
 }}
	\label{ablation}
    \centering
    \begin{adjustbox}{max width=\textwidth}
    \begin{tabular}{r|rr|rr|rr|rr|rr|rr}
\toprule
\textbf{Model} & \multicolumn{2}{c|}{\textbf{GPT3.5 (SPICE)}} & \multicolumn{2}{c|}{\textbf{GPT3.5 (w/o context)}} & \multicolumn{2}{c|}{\textbf{GPT3.5 (w/o CoT)}} & \multicolumn{2}{c|}{\textbf{GPT3.5 (w/o flow)}} & \multicolumn{2}{c|}{\textbf{GPT3.5}} & \multicolumn{2}{c}{\textbf{GPT3.5 (fine-tune)}} \\
Task ID & Pass@1 & Pass@5 & Pass@1 & Pass@5 & Pass@1 & Pass@5 & Pass@1 & Pass@5 & Pass@1 & Pass@5 & Pass@1 & Pass@5 \\
\midrule
1  & 50.0 & 97.9 & 10.0 & 43.3 & \textbf{100.0} & \textbf{100.0} & 70.0 & 99.9 & 86.7 & \textbf{100.0} & \textbf{100.0} & \textbf{100.0} \\
2  & 46.7 & 96.9 & 3.3 & 16.7 & \textbf{93.3} & \textbf{100.0} & 70.0 & 99.9 & 70.0 & 99.9 & 86.7 & \textbf{100.0} \\
3  & 0.0 & 0.0 & 0.0 & 0.0 & 0.0 & 0.0 & 0.0 & 0.0 & 3.3 & 16.7 & \textbf{40.0} & \textbf{95.8} \\
4  & 23.3 & 76.4 & 26.7 & 81.5 & 0.0 & 0.0 & 46.7 & 96.9 & 50.0 & 97.9 & \textbf{80.0} & \textbf{100.0} \\
5  & 10.0 & 43.3 & 0.0 & 0.0 & 3.3 & 16.7 & 6.7 & 31.0 & 10.0 & 43.3 & \textbf{20.0} & \textbf{73.6} \\
6  & 53.3 & 98.6 & 86.7 & \textbf{100.0} & 83.3 & \textbf{100.0} & 53.3 & 98.6 & 73.3 & \textbf{100.0} & \textbf{100.0} & \textbf{100.0} \\
7  & 83.3 & \textbf{100.0} & 26.7 & 81.5 & 76.7 & \textbf{100.0} & 40.0 & 94.0 & 76.7 & \textbf{100.0} & \textbf{86.7} & \textbf{100.0} \\
8  & 60.0 & 99.4 & 33.3 & 89.1 & 53.3 & 98.6 & 10.0 & 43.3 & 66.7 & 99.8 & \textbf{93.3} & \textbf{100.0} \\
9  & 3.3 & 16.7 & 0.0 & 0.0 & \textbf{53.3} & \textbf{98.6} & 0.0 & 0.0 & 30.0 & 85.7 & 26.7 & 84.6 \\
10 & 3.3 & 16.7 & 6.7 & 31.0 & 3.3 & 16.7 & 10.0 & 43.3 & \textbf{46.7} & \textbf{96.9} & 40.0 & 95.8 \\
11-24 & 0.0 & 0.0 & 0.0 & 0.0 & 0.0 & 0.0 & 0.0 & 0.0 & 0.0 & 0.0 & 0.0 & 0.0  \\
\midrule
Avg & 13.9 & 26.9 & 8.1 & 18.5 & 19.4 & 26.3 & 12.8 & 25.3 & 21.4 & 35.0 & \textbf{28.1} & \textbf{39.6} \\
\# Solved & 9 & 9 & 7 & 7 & 8 & 8 & 8 & 8 & \textbf{10} & \textbf{10} & \textbf{10} & \textbf{10} \\
\bottomrule
\end{tabular}
    \end{adjustbox}
\vspace{-10pt}
\end{table}

\paragraph{Fine-tuning.} 
We employed a 3-fold cross-validation for fine-tuning evaluation, using two subsets of design tasks for fine-tuning and the remaining one for testing. 
Fine-tuning was conducted using the API of GPT-3.5 with two epochs. The results are shown in Table \ref{ablation}. 
Fine-tuned GPT-3.5 generally performs better on design tasks, as fine-tuning helps standardize design outputs through correct examples and reduces common syntax and design errors. 
However, due to the inherent limitations of the GPT-3.5 base model, fine-tuned models struggle to design additional circuits when data is limited.

\begin{figure}[!t]
  \centering
  \includegraphics[width=\textwidth]{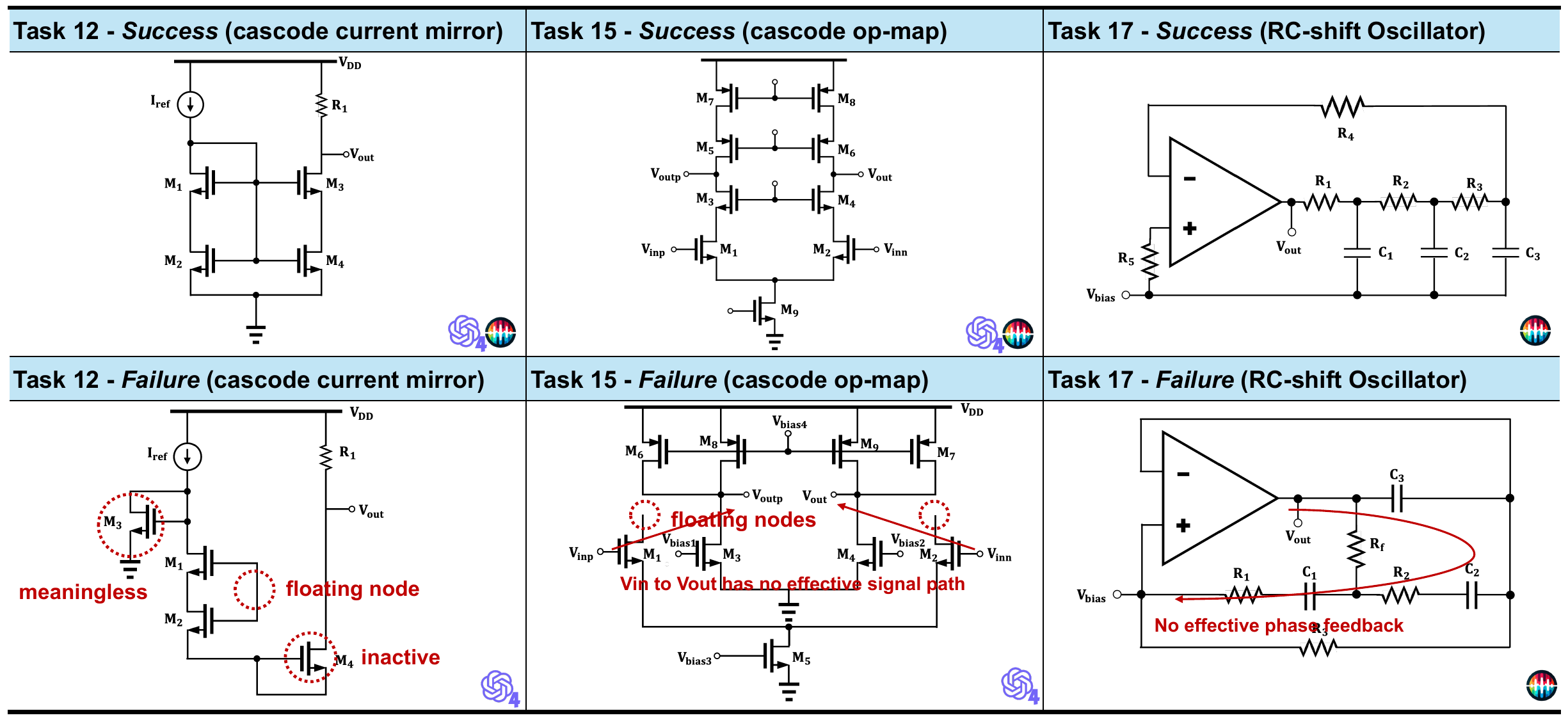}
  \caption{\small{\textbf{Visualization for successful and failed designs.} 
  The LLM model source utilized for this circuit's design is detailed in the lower right corner.
  }
}
  \label{visualization}
  \vspace{-10pt}
\end{figure}

\paragraph{Visualization.} 
Several successful and failed circuit design diagrams are in Fig.~\ref{visualization}, with icons at the bottom of the figures indicating the corresponding source LLMs. 
It can be observed that even the slightest discrepancy can render a circuit non-functional. 
More results are shown in Appendix Fig.~\ref{visualization_app}.

\begin{wrapfigure}{r}{0.5\textwidth}
    \vspace{-10pt}
  \centering
  \includegraphics[width=0.49\textwidth]{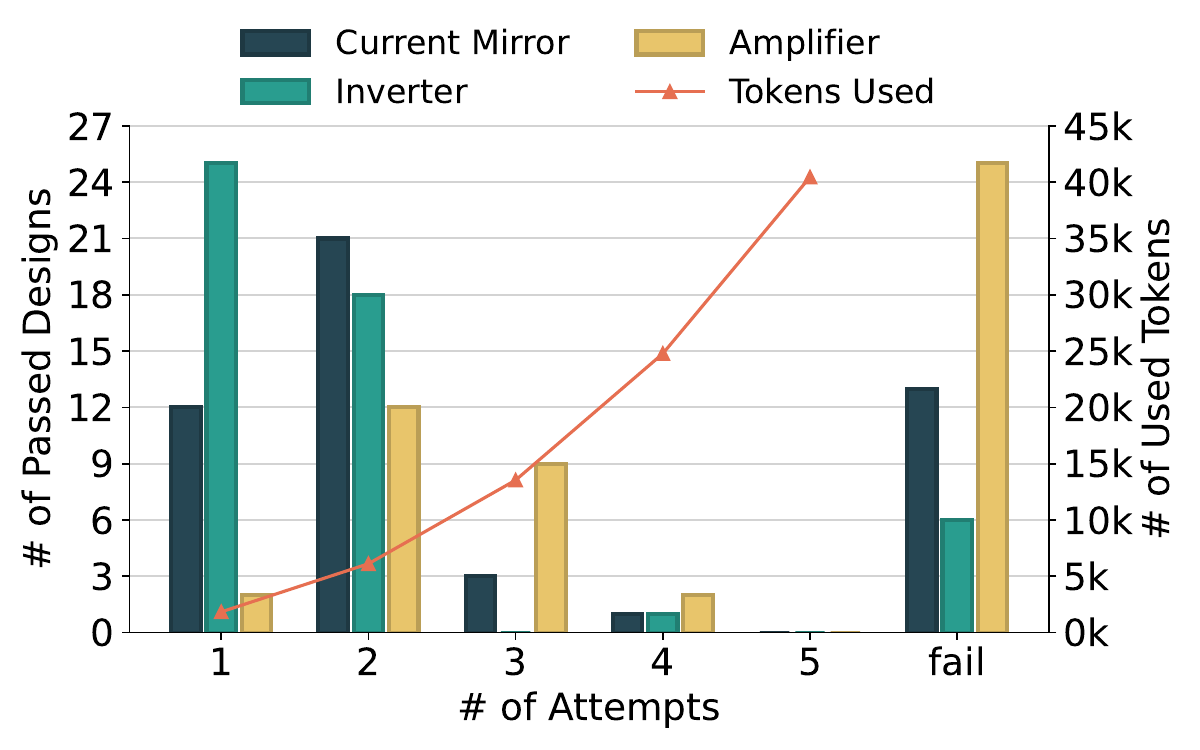}
  \caption{\small{\textbf{Attempt Times.} 50 trials for each task with a maximum of 5 attempts across design tasks, the findings indicate that success probability significantly drops after the third attempt.
  }}
  \label{try_times}
  \vspace{-10pt}
\end{wrapfigure}

\paragraph{Attempt Times.} 
The number of attempts is a hyper-parameter to maximize the benefit-cost ratio. 
We tested three tasks (Task ID=7, 8, 10) and conducted 50 trials with a maximum of 5 attempts per trial using GPT-3.5. 
Fig.~\ref{try_times} shows the distribution of successful design attempts across 50 trials.
Results show that most designs are completed within three attempts. 
If a design is not achieved within this limit, the LLM's further attempts are unlikely to succeed, and token consumption continues to rise. Consequently, we have set the default number of attempts to three. 
The number of design attempts for composite circuits has been limited to two because they involve more complex and challenging fixes and have high generation costs.

\section{Conclusion}

This paper proposes AnalogCoder, a training-free LLM agent for automatic analog circuit design. It innovatively transforms the analog circuit design tasks into the generation of Python code, significantly reducing the complexity faced by LLMs. At the same time, it is equipped with crafted prompts, a feedback-enhanced design flow, and a circuit tool library, effectively enhancing the success rate of the designs. 
Also, we provide an open-source analog design benchmark for future research. 
This work facilitates the complex, time-consuming, and error-prone process of analog circuit design, enabling individuals with limited design experience to easily create analog circuits.
\looseness=-1

\paragraph{Limitation and Societal Impact.} 
Currently, LLMs lack the capability to design highly complex analog circuits, but future advancements may address this.
Additionally, their use is constrained by costs and the availability of computational resources. 
As LLMs improve, there is a risk of AI replacing parts of human roles, a challenge common to all projects involving LLMs.

{
\small

\bibliography{references}

}

\clearpage

\appendix
\section{Appendix}

\subsection{Benchmark Details}
\label{sec:benchmark_detail}

We list all details of the analog circuit design benchmark details in Table \ref{benchmark_detail}.

\begin{table}[ht]
\centering
\caption{Benchmark Details}
\label{benchmark_detail}
\resizebox{\textwidth}{!}{
\begin{tabular}{c|l|L{8cm}|L{3.3cm}|l|L{1.6cm}}
\toprule
\textbf{Id} & \textbf{Type} & \textbf{Design Task Description} & \textbf{Input} & \textbf{Output} & \textbf{Composite Circuit} \\ \midrule
\rowcolor{lightGreen}
1 & Amplifier & a single-stage common-source amplifier with resistive load R & Vin & Vout & No \\
\rowcolor{lightGreen}
2 & Amplifier & a three-stage amplifier with single input and output (each stage is common-source with resistive load) & Vin & Vout & No \\
\rowcolor{lightGreen}
3 & Amplifier & a common-drain amplifier (a.k.a. a source follower) with resistive load R (output Vout at the source) & Vin & Vout & No \\
\rowcolor{lightGreen}
4 & Amplifier & a single-stage common-gate amplifier with resistive load R (input signal Vin must be applied at the source terminal) & Vin, Vbias & Vout & No \\
\rowcolor{lightGreen}
5 & Amplifier & a single-stage cascode amplifier with two NMOS transistors provides a single-ended output through a resistive load R & Vin, Vbias & Vout & No \\
\rowcolor{lightGreen}
6 & Inverter & a NMOS inverter with resistive load R & Vin & Vout & No \\
\rowcolor{lightGreen}
7 & Inverter & a logical inverter with 1 NMOS and 1 PMOS & Vin & Vout & No \\
\rowcolor{lightGreen}
8 & CurrentMirror & a simple NMOS constant current source with resistive load R & Vbias & Vout & No \\ 
\rowcolor{lightOrange}
9 & Amplifier & a single-stage amplifier (common-source with PMOS diode-connected load (gate and drain are shorted)) & Vin & Vout & No \\
\rowcolor{lightOrange}
10 & Amplifier & a two-stage amplifier with a Miller compensation capacitor & Vin & Vout & No \\
\rowcolor{lightOrange}
11 & Opamp & a differential opamp with an active PMOS current mirror load, a tail current source, and two outputs & Vinp, Vinn, Vbias & Voutp, Vout & No \\
\rowcolor{lightOrange}
12 & CurrentMirror & A cascode current mirror with 4 mosfets (2 stacked at input side with diode-connected, 2 stacked at output side), reference current source input Iref (connected to Vdd) and resistive load R & Iref & Iout & No \\ 
\rowcolor{lightOrange}
13 & Opamp & a single-stage differential common-source opamp with dual resistive loads, tail current, and a single output & Vinp, Vinn & Vout & No \\
\rowcolor{lightRed}
14 & Opamp & a two-stage differential opamp (first stage: common-source with an active load and a tail current, second stage: common-source with an active load) & Vinp, Vinn, Vbias1, Vbias2, Vbias3 & Voutp, Vout & No \\
\rowcolor{lightRed}
15 & Opamp & a single-stage telescopic cascode opamp with two outputs (4 nmos as cascode input pair, 4 pmos as cascode loads, and 1 tail current) & Vinp, Vinn, Vbias1, Vbias2, Vbias3, Vbias4 & Voutp, Vout & No \\
\rowcolor{lightRed}
16 & Oscillator & an RC phase-shift oscillator & - & Vout & Yes \\
\rowcolor{lightRed}
17 & Oscillator & a Wien Bridge oscillator & - & Vout & Yes \\
\rowcolor{lightRed}
18 & Integrator & an Opamp integrator with resistor R1 and capacitor Cf & Vin & Vout & Yes \\
\rowcolor{lightRed}
19 & Differentiator & an Opamp differentiator with resistor Rf and capacitor C1 & Vin & Vout & Yes \\
\rowcolor{lightRed}
20 & Adder & an Opamp adder to make Vout=-(Vin1+Vin2) & Vin1, Vin2 & Vout & Yes \\
\rowcolor{lightRed}
21 & Subtractor & an Op-amp subtractor to make Vout=Vin2-Vin1 & Vin1, Vin2 & Vout & Yes \\
\rowcolor{lightRed}
22 & Schmitt Trigger & a non-inverting Schmitt trigger with positive feedback op-amp & Vin & Vout & Yes \\
\rowcolor{lightRed}
23 & VCO & a voltage-controlled oscillator & Vin & Vout & Yes \\
\rowcolor{lightRed}
24 & PLL & a phase-locked loop & CLK\textsubscript{ref} & CLK\textsubscript{p} & Yes \\
\bottomrule
\end{tabular}
}
\end{table}

The functional testing criteria for all categories of analog circuit design are enumerated in Table \ref{criteria}.

\begin{table}[ht]
\centering
\caption{\textbf{Function Correctness Criteria}}
\label{criteria}
\resizebox{\textwidth}{!}{
\begin{tabular}{c|L{12cm}}
\toprule
\textbf{Type} & \textbf{Criteria for Functional Correctness} \\ \midrule
Amplifier & Gain $A_v > 0$; Drain Current $I_D > 0$ \\ \midrule
CurrentMirror & Drain Current $I_D > 0$; Adjust the load resistor $R$ from $100 \Omega$ to $1k \Omega$, $\delta I_D < 1 \times 10^{-5}$ for at least one interval. \\ \midrule
Inverter &  $V_{out} \leq 2.5$ when $V_{in}=0$; $V_{out} \geq 2.5$ when $V_{in}=V_{dd}$; $\delta V_{out} \leq 1.0V$. \\ \midrule
Opamp &   Drain Current $I_D > 0$; Differential-mode gain $A_{DM} >0$ when $f=100$Hz; Differential-mode gain greater than common-mode gain $A_{DM}>A_{CM}$ when $f=100$Hz.\\ \midrule
Oscillator &  Within 10 ms analysis, Number of oscillation peaks: \( N > 3 \); Oscillation amplitude: \( A > 1 \times 10^{-6} \); Oscillation period variability: \( \frac{\Delta T}{T} \leq 20\% \).\\ \midrule
Integrator &  Upon inputting a square wave, ensure that the slope \( k \) of the resulting triangular wave satisfies \( |k - k'|/k' \leq 0.3 \), where \( k' \) is derived from the standard RC time constant; The linear fit of the triangular wave's rising edge should have \( R^2 > 0.9 \); The number of peaks in the output triangular wave $>2$; The circuit configuration must not constitute a passive integrator. \\ \midrule
Differentiator &  Upon inputting a triangular wave, ensure that the number of peaks \( N > 0 \); The symmetry of the square wave's peaks and troughs relative to the base voltage \( V_0 \) should be maintained, i.e., if \( V_{peak} \) and \( V_{trough} \) represent the peak and trough voltages respectively, their deviations from \( V_0 \) should satisfy \( |V_{peak} - V_0| = |V_0 - V_{trough}| \); The fidelity of the square waveform should be such that the measured peak and trough voltages reach at least 90\% of their expected values; The circuit configuration must not constitute a passive differentiator.\\ \midrule
Adder &  Upon sweeping \( V_{in1} \) from the base voltage \( V_0 \) to \( V_0 + 0.5V \), ensure that the error between the output voltage \( V_{out} \) and the negative sum of the inputs \(-(V_{in1} + V_{in2})\) remains within 20\%, formally, the error \( \epsilon \) is defined as:
\[ \epsilon = \left|\frac{V_{out} + (V_{in1} + V_{in2})}{V_{in1} + V_{in2}}\right| \leq 0.2 \] \\ \midrule
Subtractor &  Upon sweeping \( V_{in1} \) from \( 2V_0 - 2.25 \) to \( 2V_0 - 1.75 \), with \( V_{in2} \) set at \( 2V_0 \) ($V_0$ is bias voltage), ensure that the error between the output voltage \( V_{out} \) and the difference between \( V_{in2} \) and \( V_{in1} \) (\( V_{in2} - V_{in1} \)) remains within 20\%, formally, the error \( \epsilon \) is defined as:
\[ \epsilon = \left| \frac{V_{out} - (V_{in2} - V_{in1})}{V_{in2} - V_{in1}} \right| \leq 0.2 \]
\\ \midrule
Schmitt Trigger & Ensure \( V_{out} \) crosses \( V_{dd}/2 \) at least once when \( V_{in} \) is swept from 0 to \( V_{dd} \) and back to 0; The difference in \( V_{in} \) at which \( V_{out} \) reaches \( V_{dd}/2 \) during the upward and downward sweeps should exceed 0.05V, formally \( |V_{in, high} - V_{in, low}| > 0.05 \); \( V_{out} \) should be a monotonic function of \( V_{in} \) throughout each sweep. \\ \midrule
VCO & Applying $V_{in}$ of 0.7V, 0.8V, and 0.85V, measure the corresponding output periods \( T_{out} \) of $V_{out}$, denoted as \( T_{0.7} \), \( T_{0.8} \), and \( T_{0.85} \), and satisfying either  \( \left| T_{0.7} - T_{0.8} \right| > \epsilon \, \text{and} \, \left| T_{0.8} - T_{0.85} \right| > \epsilon \, \text{and} \, \left| T_{0.7} - T_{0.85} \right| > \epsilon. \) , \( \text{where} \, \epsilon = 10^{-6} \) . \\ \midrule
PLL &  Apply a 10 MHz clock to \( CLK_{ref} \) and observe whether the output frequency of \( CLK_{p} \) deviates from 10 MHz by no more than 5\%:
\[
\left| \frac{f_{CLK_p} - f_{CLK_{ref}}}{f_{CLK_{ref}}} \right| \leq 0.05
\]
\\
\bottomrule
\end{tabular}
}
\end{table}

\newpage

\subsection{Complete Prompts}
\label{complete_prompt}
We provide the complete prompts as follows. 

\begin{tcolorbox}[title={User}]
\footnotesize
You are an analog integrated circuits expert.

\end{tcolorbox}

\begin{tcolorbox}[breakable, title=Assistant (Design prompt for basic circuits)]
\footnotesize
You aim to design a topology for a given circuit described in the text. 

Please ensure your designed circuit topology works properly and achieves the design requirements. 

Here is an example:

\section*{Question}

Design a 2-stage amplifier (first stage: a common-source stage with current-source load, second stage: a common-source stage with resistor load).

Input node name: Vin, Vbias.

Output node name: Vout.

\section*{Answer}

\subsection*{Task 1}

\subsubsection*{Components Needed}

\begin{itemize}
  \item \textbf{NMOS Transistors}: M1 and M3
  \item \textbf{PMOS Transistors}: M2 (used as the current source in the first stage)
  \item \textbf{Resistors}: R1 for the second stage load
  \item \textbf{Power Supply}: Vdd for DC supply
  \item \textbf{Input Signal Source}: Vin, Vbias for biasing and signal input
  \item \textbf{Capacitors}: Not specified but can be included for coupling and bypass applications if required
\end{itemize}

\subsubsection*{Stage 1: Common-Source Amplifier with Current Source Load}

\begin{enumerate}
  \item \textbf{Transistor Setup}:
  \begin{itemize}
    \item \textbf{M1} (NMOS) as the main amplifying transistor.
    \item Gate of \textbf{M1} is connected to the input node \textbf{Vin}.
    \item Source of \textbf{M1} connected to the ground.
    \item Drain of \textbf{M1} connected to the drain of \textbf{M2}.
  \end{itemize}

  \item \textbf{Biasing}:
  \begin{itemize}
    \item \textbf{Vin} provides the input signal.
    \item \textbf{Vbias} is used to bias \textbf{M2} (PMOS), ensuring it operates as a current source.
  \end{itemize}

  \item \textbf{Current Source Load (M2)}:
  \begin{itemize}
    \item \textbf{M2}, a PMOS transistor, is configured as a current source.
    \item The source of \textbf{M2} is connected to \textbf{Vdd}, and its gate is connected to \textbf{Vbias}.
    \item Drain of \textbf{M2} is connected to the drain of \textbf{M1}, providing a high-impedance load.
  \end{itemize}
\end{enumerate}

\subsubsection*{Stage 2: Common-Source Amplifier with Resistor Load}
\begin{enumerate}
  \item \textbf{Transistor Setup}:
  \begin{itemize}
    \item \textbf{M3} (NMOS) as the main amplifying transistor for the second stage.
    \item Gate of \textbf{M3} connected to the drain of \textbf{M1}.
    \item Source of \textbf{M3} connected to the ground.
    \item Drain of \textbf{M3} connected to \textbf{Vout} through resistor \textbf{R1}.
  \end{itemize}

  \item \textbf{Load and Coupling}:
  \begin{itemize}
    \item \textbf{R1} connects the drain of \textbf{M3} to \textbf{Vdd}. This resistor converts the current through \textbf{M3} into an output voltage.
  \end{itemize}
\end{enumerate}

\subsection*{Task 2}

\begin{verbatim}
from PySpice.Spice.Netlist import Circuit
from PySpice.Unit import *

circuit = Circuit('Two-Stage Amplifier')
# Define the MOSFET models
circuit.model('nmos_model', 'nmos', level=1, kp=100e-6, vto=0.5)
circuit.model('pmos_model', 'pmos', level=1, kp=50e-6, vto=-0.5)

# Power Supplies
circuit.V('dd', 'Vdd', circuit.gnd, 5.0) # 5V power supply
circuit.V('in', 'Vin', circuit.gnd, 1.0) # 1V input for bias voltage
circuit.V('bias', 'Vbias', circuit.gnd, 4.0) # 4V input for bias voltage

# First Stage: Common-Source with Active Load
circuit.MOSFET('1', 'Drain1', 'Vin', circuit.gnd, circuit.gnd, 
model='nmos_model', w=50e-6, l=1e-6)
circuit.MOSFET('2', 'Drain1', 'Vbias', 'Vdd', 'Vdd', 
model='pmos_model', w=100e-6, l=1e-6)

# Second Stage: Common-Source with Resistor Load
circuit.MOSFET('3', 'Vout', 'Drain1', circuit.gnd, 
circuit.gnd, model='nmos_model', w=100e-6, l=1e-6)
circuit.R('1', 'Vout', 'Vdd', 1@u_kOhm) 
# Analysis Part
simulator = circuit.simulator()
\end{verbatim}

As you have seen, the output of your designed topology should consist of two tasks:
\begin{enumerate}
    \item Give a detailed design plan about all devices and their interconnectivity nodes and properties.
    \item Write a complete Python code, describing the topology of integrated analog circuits according to the design plan.
\end{enumerate}

Please make sure your Python code is compatible with PySpice.\\
Please give the runnable code without any placeholders.

Do not write other redundant codes after \texttt{simulator = circuit.simulator()}.

\section*{Tips}
\noindent There are some tips you should remember all the time:
\begin{enumerate}
    \item For the MOSFET definition \texttt{circuit.MOSFET(name, drain, gate, source, bulk, model, w=w1, l=l1)}, be careful about the parameter sequence.
    \item You should connect the bulk of a MOSFET to its source.
    \item Please use the MOSFET threshold voltage, when setting the bias voltage.
    \item Avoid giving any AC voltage in the sources, just consider the operating points.
    \item Make sure the input and output node names appear in the circuit.
    \item Avoid using subcircuits.
    \item Use nominal transistor sizing.
    \item Assume the Vdd = 5.0 V.
\end{enumerate}

\section*{Question}
\noindent Design \textcolor{orange}{[TASK]}.

\noindent Input node name: \textcolor{orange}{[INPUT]}.

\noindent Output node name: \textcolor{orange}{[OUTPUT]}.

\section*{Answer}

\end{tcolorbox}

\begin{tcolorbox}[breakable, title=Assistant (Retrieval prompt)]
\footnotesize
I have the following implemented subcircuits you can directly call them by Python code with the Pyspice library, we list all basic information on them.

\medskip
\textcolor{blue}{[TABLE]}

\medskip
Now, you need to design \textcolor{orange}{[TASK]}. Please maximize the design success rate, thus the circuit with two inputs and the highest possible gain has greater flexibility. Please choose the subcircuits from the above table you will use and make the number of chosen subcircuits as few as possible.

\medskip
Please give out the IDs of the subcircuits that you choose, and enumerate them in a Python list like \texttt{[0]}.

\end{tcolorbox}

\begin{tcolorbox}[breakable, title=Assistant (Design prompt for composite circuits)]
\footnotesize
You aim to design a topology for a given circuit described in the text. 
Please ensure your designed circuit topology works properly and achieves the design requirements. 
To make the task easier, I provide you with some existing subcircuits you can directly use by calling functions in Python with the PySpice library.
Now I would like you to help me design a complex analog circuit based on them.

Here is an example:

\section*{Question}
Design an opamp with 470 ohm resistance load.

Input node name: in

Output node name: out

You can directly use the following subcircuits.

\subsection*{Subcircuits Info}

{
\scriptsize
\begin{tabular}{ccccccc}
\textbf{Id} & \textbf{Circuit Type} & \textbf{Gain/Differential-mode gain} & \textbf{Common-mode gain} & \textbf{Input} & \textbf{Output} \\
- & Opamp & $10.00 \times 10^{0}$ & $0.00 \times 10^{0}$ & Vin & Vout \\
\end{tabular}
}

\subsection*{Call Info}

To use them, please insert the following codes.

\begin{verbatim}

from example_lib import *
# declare the subcircuit
circuit.subcircuit(‘BasicOperationalAmplifier())
# create a subcircuit instance
circuit.X('1', 'BasicOperationalAmplifier', 'Vin', 'Vout')
\end{verbatim}

\section*{Answer}

\begin{verbatim}

from PySpice.Spice.Netlist import Circuit
from PySpice.Unit import *
from example_lib import *

circuit = Circuit('Operational Amplifier')

# Define the MOSFET models
circuit.model('nmos_model', 'nmos', level=1, kp=100e-6, vto=0.5)
circuit.model('pmos_model', 'pmos', level=1, kp=50e-6, vto=-0.5)


circuit.V('input', 'in', circuit.gnd, 2.5@u_V)
circuit.subcircuit(BasicOperationalAmplifier())
circuit.X('op', 'BasicOperationalAmplifier', 'in', circuit.gnd, 'out')
R = 470
circuit.R('load', 'out', circuit.gnd, R)

simulator = circuit.simulator()
\end{verbatim}

As you have seen, the output of your designed topology should be in a complete Python code, describing the topology of integrated analog circuits according to the design plan. 

Please make sure your Python code is compatible with PySpice. 
Please give the runnable code without any placeholders.
Do not write other redundant codes after \texttt{simulator = circuit.simulator()}.

\begin{enumerate}
    \item For the MOSFET definition circuit: \texttt{MOSFET(name, drain, gate, source, bulk, model, w=w1, l=l1)}, be careful about the parameter sequence.
    \item You should connect the bulk of a MOSFET to its source.
    \item Please use the MOSFET threshold voltage when setting the bias voltage.
    \item Avoid giving any AC voltage in the sources, just consider the operating points.
    \item Make sure the input and output node names appear in the circuit.
    \item Assume the Vdd = 5.0 V. Do not need to add the power supply for subcircuits.
\end{enumerate}

\section*{Question}

Design \textcolor{orange}{[TASK]}. 

Input node name: \textcolor{orange}{[INPUT]}.

Output node name: \textcolor{orange}{[OUTPUT]}.

You can directly use the following subcircuits.

\subsection*{Subcircuits Info}

\textcolor{purple}{[SUBCIRCUITS\_INFO]}

\subsection*{Note}

\textcolor{purple}{[NOTE\_INFO]}

\subsection*{Call Info}

To use them, please insert the following codes.

\textcolor{purple}{[CALL\_INFO]}

\section*{Answer}

\end{tcolorbox}

In the prompts, the sections enclosed in square brackets serve as placeholders and will be replaced with specific content pertinent to the designated task.
The placeholders \textcolor{orange}{[TASK]}, \textcolor{orange}{[INPUT]}, and \textcolor{orange}{[OUTPUT]} are replaced with information regarding the design of circuits as specified in Table \ref{benchmark_detail}.

The placeholder \textcolor{blue}{[TABLE]} is used to display key information relevant to the current circuit tool library, with a specific example provided in Table \ref{circuit_library_overview}.

The placeholder \textcolor{purple}{[SUBCIRCUITS\_INFO]} gives the basic info selected subcircuits' information in one table. An example is as follows.

\begin{tcolorbox}[breakable, title=SUBCIRCUITS\_INFO]
{
\scriptsize
\begin{tabular}{ccccccc}
\textbf{Id} & \textbf{Circuit Type} & \textbf{Gain/Differential-mode gain} & \textbf{Common-mode gain} & \textbf{Input} & \textbf{Output} \\
11 & Opamp & 193.98 & -173.70 & Vinp, Vinn & Vout \\
\end{tabular}
}
\end{tcolorbox}

The placeholder \textcolor{purple}{[NOTE\_INFO]} gives supplementary information for using subcircuits, which is built based on the circuit tool library as shown in Table~\ref{circuit_library_overview} and the text template. An example is as follows, wherein the \textit{SingleStageOpamp} is the function name.

\begin{tcolorbox}[breakable, title=NOTE\_INFO]
\footnotesize
The Vinn of SingleStageOpamp is the inverting input.

The Vinp of SingleStageOpamp is the non-inverting input.

The DC operating voltage for Vinn/Vinp is 2.5 V.

\medskip
\textit{(only for the Oscillator circuit)}

Due to the operational range of the op-amp being 0 to 5V, please connect the nodes that were originally grounded to a 2.5V DC power source.

Please increase the gain as much as possible to maintain oscillation.
\end{tcolorbox}

The placeholder \textcolor{purple}{[CALL\_INFO]} provides the function invocation methods when using the subcircuits. An example is as follows.

\begin{tcolorbox}[breakable, title=CALL\_INFO]
\footnotesize
To use them, please insert the following codes.

\begin{verbatim}
from p11_lib import *
# declare the subcircuit
circuit.subcircuit(SingleStageOpamp())
# create a subcircuit instance
circuit.X('1', 'SingleStageOpamp', 'Vinp', 'Vinn', 'Vout')
\end{verbatim}
\end{tcolorbox}

The implement of the subcircuit \textit{SingleStageOpamp} is saved in the circuit tool library as values as shown in Fig.~\ref{skill_library}, which is automatically transformed into the SubCircuit class through scripting. An example of the implementation is as follows.

\begin{minted}
[
frame=lines,
framesep=2mm,
baselinestretch=1.2,
fontsize=\scriptsize,
linenos
]
{python}
from PySpice.Unit import *
from PySpice.Spice.Netlist import SubCircuitFactory
class SingleStageOpamp(SubCircuitFactory):
    NAME = ('SingleStageOpamp')
    NODES = ('Vinp', 'Vinn', 'Vout')
    def __init__(self):
        super().__init__()
        # Define the MOSFET models
        self.model('nmos_model', 'nmos', level=1, kp=100e-6, vto=0.5)
        self.model('pmos_model', 'pmos', level=1, kp=50e-6, vto=-0.5)
        # Power Supplies
        self.V('dd', 'Vdd', self.gnd, 5.0)  # 5V power supply
        self.V('bias', 'Vbias', self.gnd, 1.5)  # Bias voltage for the tail current source M3
        # Input Voltage Sources for Differential Inputs
        # Differential Pair and Tail Current Source
        self.MOSFET('1', 'Voutp', 'Vinp', 'Source3', 'Source3', model='nmos_model', w=50e-6, l=1e-6)
        self.MOSFET('2', 'Vout', 'Vinn', 'Source3', 'Source3', model='nmos_model', w=50e-6, l=1e-6)
        self.MOSFET('3', 'Source3', 'Vbias', self.gnd, self.gnd, model='nmos_model', w=100e-6, l=1e-6)
        # Active Current Mirror Load
        self.MOSFET('4', 'Voutp', 'Voutp', 'Vdd', 'Vdd', model='pmos_model', w=100e-6, l=1e-6)
        self.MOSFET('5', 'Vout', 'Voutp', 'Vdd', 'Vdd', model='pmos_model', w=100e-6, l=1e-6)
\end{minted}

We also give the prompts of `GPT-3.5 (SPICE)', `GPT-3.5 (w/o context)', and `GPT-3.5 (w/o CoT)' in our ablation studies. 
`GPT-3.5 (SPICE)' makes the LLMs generate SPICE code rather than Python code, whereas NgSPICE is a widely-used open-source version of SPICE. 
`GPT-3.5 (w/o context)' disregards the context example in the prompt, meaning that a demonstrative design question and answer should not be provided. 
`GPT-3.5 (w/o CoT)' omits the Chain-of-Thought component, meaning it allows the language model to generate code directly without first enumerating the required components and their connections.
We have omitted portions of the content identical to the original prompt.

\begin{tcolorbox}[breakable, title=Assistant (Design prompt for basic circuits 
 / SPICE)]
\footnotesize
... \textit{(same with Assistant (Design prompt for basic circuits))}
\subsection*{Task 2}

\begin{verbatim}
* Two-Stage Amplifier

* Define the MOSFET models
.model nmos_model nmos level=1 kp=100e-6 vto=0.5
.model pmos_model pmos level=1 kp=50e-6 vto=-0.5

* Power Supplies for the power and input signal
Vdd Vdd 0 5.0
Vin Vin 0 1.0
Vbias Vbias 0 4.0

* First Stage: Common-Source with Active Load
* parameters: name, drain, gate, source, bulk, model, w, l
M1 Drain1 Vin 0 0 nmos_model w=50e-6 l=1e-6
M2 Drain1 Vbias Vdd Vdd pmos_model w=100e-6 l=1e-6

* Second Stage: Common-Source with Resistor Load
M3 Vout Drain1 0 0 nmos_model w=100e-6 l=1e-6
R1 Vout Vdd 1k

.end
\end{verbatim}

As you have seen, the output of your designed topology should consist of two tasks:

\begin{enumerate}
\item Give a detailed design plan about all devices and their interconnectivity nodes and properties.
\item Write a complete \textbf{NgSpice} code, describing the topology of integrated analog circuits according to the design plan. 
\end{enumerate}

Please give the runnable code without any placeholders.

Do not write other redundant codes after \textbf{\texttt{.end}}.

\medskip
... \textit{(same with Assistant (Design prompt for basic circuits))}
\end{tcolorbox}

\begin{tcolorbox}[breakable, title=Assistant (Design prompt for basic circuits / without in-context learning)]
\footnotesize
You aim to design a topology for a given circuit described in the text. 

Please ensure your designed circuit topology works properly and achieves the design requirements. 

The output of your designed topology should consist of two tasks:
\begin{enumerate}
\item Give a detailed design plan about all devices and their interconnectivity nodes and properties.
\item Write a complete Python code, describing the topology of integrated analog circuits according to the design plan. 
\end{enumerate}

Please make sure your Python code is compatible with PySpice. 

Please give the runnable code without any placeholders.

Do not write other redundant codes after \texttt{simulator = circuit.simulator()}.

For importing libraries, you can use:
\begin{verbatim}
from PySpice.Spice.Netlist import Circuit
from PySpice.Unit import *
\end{verbatim}

For the mosfet, you can refer to the following code:
\begin{verbatim}
circuit.model('nmos_model', 'nmos', level=1, kp=100e-6, vto=0.5)
circuit.model('pmos_model', 'pmos', level=1, kp=50e-6, vto=-0.5)
circuit.MOSFET('1', 'Vout', 'Vin', circuit.gnd, circuit.gnd, 
model='nmos_model', w=50e-6, l=1e-6)
\end{verbatim}

For the resistor and the voltage source, you can can refer to the following code:
\begin{verbatim}
circuit.R('1', 'Vout', 'Vdd', 1000)
circuit.V('dd', 'Vdd', circuit.gnd, 5.0)
\end{verbatim}

There are some tips you should remember all the time:

\medskip
... \textit{(same with Assistant (Design prompt for basic circuits))}

\end{tcolorbox}

\begin{tcolorbox}[breakable, title=Assistant (Design prompt for basic circuits / without CoT)]
You aim to design a topology for a given circuit described in the text. 

Please ensure your designed circuit topology works properly and achieves the design requirements. 

Here is an example:

\section*{Question}

Design a 2-stage amplifier (first stage: a common-source stage with current-source load, second stage: a common-source stage with resistor load).

Input node name: Vin, Vbias.

Output node name: Vout.

\section*{Answer}
\footnotesize
\begin{verbatim}
from PySpice.Spice.Netlist import Circuit
from PySpice.Unit import *

circuit = Circuit('Two-Stage Amplifier')
# Define the MOSFET models
circuit.model('nmos_model', 'nmos', level=1, kp=100e-6, vto=0.5)
circuit.model('pmos_model', 'pmos', level=1, kp=50e-6, vto=-0.5)

# Power Supplies
circuit.V('dd', 'Vdd', circuit.gnd, 5.0) # 5V power supply
circuit.V('in', 'Vin', circuit.gnd, 1.0) # 1V input for bias voltage
circuit.V('bias', 'Vbias', circuit.gnd, 4.0) # 4V input for bias voltage

# First Stage: Common-Source with Active Load
circuit.MOSFET('1', 'Drain1', 'Vin', circuit.gnd, circuit.gnd, 
model='nmos_model', w=50e-6, l=1e-6)
circuit.MOSFET('2', 'Drain1', 'Vbias', 'Vdd', 'Vdd', 
model='pmos_model', w=100e-6, l=1e-6)

# Second Stage: Common-Source with Resistor Load
circuit.MOSFET('3', 'Vout', 'Drain1', circuit.gnd, 
circuit.gnd, model='nmos_model', w=100e-6, l=1e-6)
circuit.R('1', 'Vout', 'Vdd', 1@u_kOhm) 
# Analysis Part
simulator = circuit.simulator()
\end{verbatim}

As you have seen, the output of your designed topology should be in a complete Python code, describing the topology of integrated analog circuits according to the design plan. 

Please make sure your Python code is compatible with PySpice.

Please give the runnable code without any placeholders.

Do not write other redundant codes after \texttt{simulator = circuit.simulator()}.

There are some tips you should remember all the time:

\medskip
... \textit{(same with Assistant (Design prompt for basic circuits))}
\end{tcolorbox}

\subsection{Circuit Tool Library Overview}

As discussed in the circuit tool library, all successfully implemented basic circuits will be saved into the tool library. 
Information such as the attributes of the circuits will be stored in the library as keys and values. 
The key information of the tool library is listed in Table \ref{circuit_library_overview}.

\begin{table}[ht]
\caption{\textbf{Circuit Tool Library Overview}}
\label{circuit_library_overview}
\resizebox{\textwidth}{!}{
\begin{tabular}{clcccccc}
\toprule
\textbf{Task Id} & \textbf{Type} & \textbf{Gain (dB)} & \textbf{CMG (dB)\textsuperscript{*}} & \textbf{\# of inputs} & \textbf{\# of outputs} & \textbf{Input to $V_{out}$/$I_{out}$ Phase} \\ 
\midrule
1  & Amplifier    & 13.98  & NA     & 1 & 1 & inverting \\
2  & Amplifier    & 44.13  & NA     & 1 & 1 & inverting \\
3  & Amplifier    & -1.58  & NA     & 1 & 1 & non-inverting \\
4  & Amplifier    & 13.98  & NA     & 1 & 1 & non-inverting \\
5  & Amplifier    & 13.98  & NA     & 1 & 1 & inverting \\
6  & Inverter     & NA     & NA     & 1 & 1 & NA \\
7  & Inverter     & NA     & NA     & 1 & 1 & NA \\
8  & Current Mirror& NA     & NA     & 1 & 1 & NA \\
9  & Amplifier    & 75.94  & NA     & 1 & 1 & -90 degree \\
10 & Amplifier    & 6.02   & NA     & 1 & 1 & inverting \\
11 & Opamp        & 193.98 & -173.7 & 2 & 2 & non-inverting, inverting \\
12  & Current Mirror& NA     & NA     & 1 & 1 & NA \\
13 & Opamp        & 13.98  & -163.29& 2 & 1 & non-inverting, inverting \\
14 & Opamp        & -37.06 & -54.72 & 2 & 2 & non-inverting, inverting \\
15 & Opamp        & -28.15 & -72.72 & 2 & 2 & non-inverting, inverting \\
\bottomrule
\end{tabular}
}
\scriptsize
\textsuperscript{*}Common Mode Gain (CMG). The ratio by which an amplifier increases identical input signals, ideally zero in op-amps.
\end{table}

\subsection{Examples of Generated Circuits}
\label{example_generated_circuits}
We give two examples of the circuits generated by the AnalogCoder: the basic circuit (Task ID=15, a cascade op-map) and the composite circuit (Task ID=16, an RC shift Oscillator).

Here is the basic circuit example.
\begin{minted}
[
frame=lines,
framesep=2mm,
baselinestretch=1.2,
fontsize=\scriptsize,
linenos
]
{python}
from PySpice.Spice.Netlist import Circuit
from PySpice.Unit import *
circuit = Circuit('Single-Stage Telescopic Cascode Opamp')
# Define the MOSFET models
circuit.model('nmos_model', 'nmos', level=1, kp=100e-6, vto=0.5)
circuit.model('pmos_model', 'pmos', level=1, kp=50e-6, vto=-0.5)
# Power Supply
circuit.V('dd', 'Vdd', circuit.gnd, 5.0) # 5V power supply
# Input and Bias Voltages
circuit.V('inp', 'Vinp', circuit.gnd, 1.49)
circuit.V('inn', 'Vinn', circuit.gnd, 1.49)
circuit.V('bias1', 'Vbias1', circuit.gnd, 1.5@u_V) # Bias for NMOS cascode
circuit.V('bias2', 'Vbias2', circuit.gnd, 1.5@u_V) # Bias for NMOS cascode
circuit.V('bias3', 'Vbias3', circuit.gnd, 3.5@u_V) # Bias for PMOS cascode
circuit.V('bias4', 'Vbias4', circuit.gnd, 3.5@u_V) # Bias for PMOS cascode
circuit.V('biasTail', 'VbiasTail', circuit.gnd, 1.0@u_V) # Bias for the tail current source
# NMOS Transistors
circuit.MOSFET('1', 'Drain1', 'Vinp', 'Source5', circuit.gnd, model='nmos_model', w=50e-6, l=1e-6)
circuit.MOSFET('2', 'Drain2', 'Vinn', 'Source5', circuit.gnd, model='nmos_model', w=50e-6, l=1e-6)
circuit.MOSFET('3', 'Voutp', 'Vbias1', 'Drain1', circuit.gnd, model='nmos_model', w=50e-6, l=1e-6)
circuit.MOSFET('4', 'Vout', 'Vbias2', 'Drain2', circuit.gnd, model='nmos_model', w=50e-6, l=1e-6)
circuit.MOSFET('5', 'Source5', 'VbiasTail', circuit.gnd, circuit.gnd, model='nmos_model', w=50e-6, l=1e-6)
# PMOS Transistors
circuit.MOSFET('6', 'Voutp', 'Vbias3', 'Vdd', 'Vdd', model='pmos_model', w=100e-6, l=1e-6)
circuit.MOSFET('7', 'Voutp', 'Vbias4', 'Vdd', 'Vdd', model='pmos_model', w=100e-6, l=1e-6)
circuit.MOSFET('8', 'Vout', 'Vbias3', 'Vdd', 'Vdd', model='pmos_model', w=100e-6, l=1e-6)
circuit.MOSFET('9', 'Vout', 'Vbias4', 'Vdd', 'Vdd', model='pmos_model', w=100e-6, l=1e-6)
# Analysis Part
simulator = circuit.simulator()
\end{minted}

Here is the composite circuit example, which uses the subcircuit \textit{SingleStageOpamp}.

\begin{minted}
[
frame=lines,
framesep=2mm,
baselinestretch=1.2,
fontsize=\scriptsize,
linenos
]
{python}
from PySpice.Spice.Netlist import Circuit
from PySpice.Unit import *
from p11_lib import *
circuit = Circuit('RC Phase-Shift Oscillator')
# Define the power supply for non-grounded nodes
circuit.V('mid', 'mid', circuit.gnd, 2.5@u_V)  # 2.5V for biasing
# Define the operational amplifier from the library
circuit.subcircuit(SingleStageOpamp())
circuit.X('opamp', 'SingleStageOpamp', 'non_inv_input', 'inv_input', 'Vout')
# Feedback network
# RC Phase Shift Network
R_value = 10@u_kOhm
C_value = 10@u_nF
# First RC stage
circuit.R('1', 'Vout', 'n1', R_value)
circuit.C('1', 'n1', 'mid', C_value)
# Second RC stage
circuit.R('2', 'n1', 'n2', R_value)
circuit.C('2', 'n2', 'mid', C_value)
# Third RC stage
circuit.R('3', 'n2', 'inv_input', R_value)
circuit.C('3', 'inv_input', 'mid', C_value)
# Non-inverting input connected to midpoint bias
circuit.V('non_inv_input', 'non_inv_input', 'mid', 0@u_V)
simulator = circuit.simulator()
\end{minted}

\subsection{Details of Feedback-Enhanced Design Flow}

According to the related method paragraph, the feedback-enhanced design flow includes 4 main steps, each step inspects a specific part of the circuit designed by the Language Model. Table ~\ref{detail_flow} lists the specific checks conducted for each part.

\begin{table}[ht]
\centering
\caption{\textbf{Details of Feedback-Enhanced Flow}}
\label{detail_flow}
\resizebox{\textwidth}{!}{
\begin{tabular}{L{2.75cm}|L{10cm}|c}
\toprule
\textbf{Stage} & \textbf{Check Process} & \textbf{Simulation} \\ \midrule
Requirement Check & 1. Check the presence of required input and output nodes. 2. Check whether the circuit meets basic requirements (e.g., the Vin of a common-gate amplifier should be connected at the source level). & - \\ \midrule
Simulation \& Operating Point Check & 1. Check that no errors occur during the simulation, \eg floating nodes. 
2. Check each MOSFET is active: Vgs>Vth; Vds> Vgs-Vth.  & OP \\ \midrule
DC Sweep Check & 1. Check whether the output varies when the input changes from 0 to Vdd.
2. Substitute the original input voltage with the Vin that produces an output closest to Vdd/2. & DC  \\ \midrule
Requirement Check & 1. Conduct functional checks corresponding to the types of circuits. (See Table~\ref{criteria}) & AC/DC/Transient \\ \bottomrule
\end{tabular}
}
\end{table}

\subsection{Experimental Settings}
\label{experimental_settings}

\paragraph{Semiconductor Devices.}
We focus on the analog integrated circuits in this work, which means all of our circuits include the (MOSFETs).
We do not specify the MOS SPICE model to be used by the LLM, but we provide a simple example in the prompt, as demonstrated.
This is a Level-1 model~\cite{razavi2000design}, which is the simplest form of the SPICE model designed to significantly reduce the complexity of the LLM's design task by using basic approximations to simulate semiconductor device behavior.
The \textit{kp} parameter represents the transconductance parameter of the transistor, typically measured in amperes per volt squared ($A/{V^2}$).
The \textit{vto} parameter is the threshold voltage, measured in volts ($V$).

All Python codes for circuits are tested and passed in the Python $3.10$ and PySpice $1.5$ versions.

\paragraph{LLM Settings.} When employing LLMs for inference, we set temperature$=0.5$ and top\_p$=1.0$ in all models to generate varied results in each trial.

\paragraph{Fine-tuning Settings.}

We have categorized the design tasks (basic circuits) that can be completed by GPT-3.5, GPT-4, and Llama-3 into three groups. We utilized the circuits designed successfully in two of these groups as data for fine-tuning. The remaining group was used for testing purposes, with other settings identical to GPT-3.5. The task grouping is detailed in Table \ref{fine-tuning_task_grouping}.
We tested using all three models for tasks not included in the fine-tuning training set (Task ID=12, 15), and none could correctly accomplish the design.

When collecting fine-tuning data, we replace all lines with \texttt{PySpice.Unit import} with \texttt{PySpice.Unit import *} to ensure the importation of all applicable units. Concurrently, we eliminate all content containing \texttt{\@u\_V} and lines featuring \texttt{circuit.I}, due to their low frequency of occurrence in the fine-tuning dataset. As a result, the outcomes post-fine-tuning may exhibit instability.

\begin{table}[ht]
\caption{\textbf{Fine-tuning task grouping.}}
\label{fine-tuning_task_grouping}
\centering
{\footnotesize
\begin{tabular}{c|c|c|c}
\toprule
Group & A & B & C \\ \midrule
Task IDs & 1, 2, 7, 10, 11  & 3, 5, 8, 14  &  4, 6, 9, 13 \\ \bottomrule
\end{tabular}
}
\end{table}

\subsection{Supplementary Experiments}
\label{supplementray_exp}

\paragraph{Additional Results.}

We also tested the capabilities of other representative LLMs in designing analog circuits, including:
Mixtral-8×7B~\cite{jiang2024mixtral}, CodeLlama-7B, 13B, 34B~\cite{roziere2023code}, Llama2-70B~\cite{touvron2023llama}, QwenCode-7B~\cite{bai2023qwen}, DeepSeek-Coder-33B~\cite{guo2024deepseek}, and Codestral-22B~\cite{mistral2024codestral}.
The results can be seen in Table~\ref{additional_results}.

\begin{table}[!ht]
	\caption{\textbf{Additional model results.}}
	\label{additional_results}
    \centering
    \begin{adjustbox}{max width=\textwidth}
    \begin{tabular}{r|rr|rr|rr|rr|rr|rr|rr|rr}
\toprule
\textbf{Model} & \multicolumn{2}{c|}{\textbf{Mixtral-8×7B}} & \multicolumn{2}{c|}{\textbf{CodeLlama-7B}} & \multicolumn{2}{c|}{\textbf{CodeLlama-13B}} & \multicolumn{2}{c|}{\textbf{CodeLlama-34B}} & \multicolumn{2}{c|}{\textbf{Llama 2-70B}} & \multicolumn{2}{c|}{\textbf{QwenCode-7B}} & \multicolumn{2}{c|}{\textbf{DeepSeek-Coder-33B}} & \multicolumn{2}{c}{\textbf{Codestral-22B}}\\
\textbf{Task Id} & \textbf{Pass@1} & \textbf{Pass@5} & \textbf{Pass@1} & \textbf{Pass@5} & \textbf{Pass@1} & \textbf{Pass@5} & \textbf{Pass@1} & \textbf{Pass@5} & \textbf{Pass@1} & \textbf{Pass@5} & \textbf{Pass@1} & \textbf{Pass@5} & \textbf{Pass@1} & \textbf{Pass@5} & \textbf{Pass@1} & \textbf{Pass@5} \\
\midrule
1 & 6.7 & 33.3 & 23.3 & 76.4 & 0.0 & 0.0 & 23.3 & 76.4 & 76.7 & 100.0 & 0.0 & 0.0 & 60.0 & 99.4 & 93.3 & 100.0 \\
2 & 0.0 & 0.0 & 0.0 & 0.0 & 0.0 & 0.0 & 13.3 & 53.8 & 0.0 & 0.0 & 6.7 & 33.3 & 10.0 & 43.3 & 56.7 & 99.1 \\
3 & 0.0 & 0.0 & 0.0 & 0.0 & 0.0 & 0.0 & 0.0 & 0.0 & 0.0 & 0.0 & 0.0 & 0.0 & 0.0 & 0.0 & 46.7 & 96.9 \\
4 & 6.7 & 33.3 & 0.0 & 0.0 & 0.0 & 0.0 & 0.0 & 0.0 & 0.0 & 0.0 & 0.0 & 0.0 & 0.0 & 0.0 & 43.3 & 95.7 \\
5 & 0.0 & 0.0 & 0.0 & 0.0 & 10.0 & 43.3 & 3.3 & 16.7 & 0.0 & 0.0 & 6.7 & 33.3 & 0.0 & 0.0 & 13.3 & 53.8 \\
6 & 86.7 & 100.0 & 13.3 & 53.8 & 3.3 & 16.7 & 0.0 & 0.0 & 10.0 & 43.3 & 6.7 & 33.3 & 0.0 & 0.0 & 73.3 & 100.0 \\
7 & 20.0 & 73.6 & 6.7 & 31.0 & 0.0 & 0.0 & 0.0 & 0.0 & 36.7 & 91.8 & 6.7 & 33.3 & 6.7 & 31.0 & 53.3 & 98.6 \\
8 & 13.3 & 57.1 & 13.3 & 53.8 & 0.0 & 0.0 & 6.7 & 31.0 & 0.0 & 0.0 & 0.0 & 0.0 & 20.0 & 70.2 & 0.0 & 0.0 \\
9 & 0.0 & 0.0 & 0.0 & 0.0 & 0.0 & 0.0 & 0.0 & 0.0 & 0.0 & 0.0 & 0.0 & 0.0 & 0.0 & 0.0 & 0.0 & 0.0\\
10 & 0.0 & 0.0 & 0.0 & 0.0 & 0.0 & 0.0 & 0.0 & 0.0 & 0.0 & 0.0 & 0.0 & 0.0 & 0.0 & 0.0 & 13.3 & 53.8 \\
11-24 & 0.0 & 0.0 & 0.0 & 0.0 & 0.0 & 0.0 & 0.0 & 0.0 & 0.0 & 0.0 & 0.0 & 0.0 & 0.0 & 0.0 & 0.0 & 0.0 \\
\midrule
Avg & 5.6 & 12.4 & 2.4 & 9.0 & 0.6 & 2.5 & 1.9 & 7.4 & 5.1 & 9.8 & 1.1 & 5.6 & 4.0 & 10.2 & 16.4 & 29.1\\
\# Solved & 5 & 5 & 4 & 4 & 2 & 2 & 4 & 4 & 3 & 3 & 4 & 4 & 4 & 4 & 8 & 8\\
\bottomrule
\end{tabular}
    \end{adjustbox}
\end{table}

We also compared the results between the GPT-4 with circuit tool library and AnalogCoder (GPT-4o with circuit tool library), and the results are in Table~\ref{gpt4_gpt4o}. Based on the results, the capabilities of GPT-4 and GPT-4o in designing analog circuits are similar; however, GPT-4o successfully designed one additional circuit. Considering the overall costs, we have decided to implement our AnalogCoder based on GPT-4o.

\begin{table}[!ht]
	\caption{\small{\textbf{Comparasion between GPT-4 and GPT-4o.}}}
	\label{gpt4_gpt4o}
    \centering
    {\footnotesize
    \begin{tabular}{r|rr|rr}
\toprule
\textbf{Model} & \multicolumn{2}{c|}{\textbf{GPT-4 (w/ tool lib.)}} & \multicolumn{2}{c}{\textbf{AnalogCoder}} \\
Task ID & Pass@1 & Pass@5 & Pass@1 & Pass@5 \\
\midrule
1  & \textbf{100.0} & \textbf{100.0} & \textbf{100.0} & \textbf{100.0} \\
2  & \textbf{100.0} & \textbf{100.0} & \textbf{100.0} & \textbf{100.0} \\
3  & \textbf{100.0} & \textbf{100.0} & \textbf{100.0} & \textbf{100.0} \\
4  & \textbf{100.0} & \textbf{100.0} & \textbf{100.0} & \textbf{100.0} \\
5  & \textbf{100.0} & \textbf{100.0} & \textbf{100.0} & \textbf{100.0} \\
6  & \textbf{100.0} & \textbf{100.0} & \textbf{100.0} & \textbf{100.0} \\
7  & \textbf{100.0} & \textbf{100.0} & \textbf{100.0} & \textbf{100.0} \\
8  & \textbf{100.0} & \textbf{100.0} & \textbf{100.0} & \textbf{100.0} \\
9  & 73.3  & \textbf{100.0} & \textbf{100.0} & \textbf{100.0} \\
10 & \textbf{100.0} & \textbf{100.0} & \textbf{100.0} & \textbf{100.0} \\
11 & 66.7  & \textbf{100.0} & \textbf{100.0} & \textbf{100.0} \\
12 & 0.0   & 0.0   & \textbf{13.3}  & \textbf{57.1}  \\
13 & 73.3  & \textbf{100.0} & \textbf{100.0} & \textbf{100.0} \\
14 & \textbf{86.7}  & \textbf{100.0} & 73.3  & \textbf{100.0} \\
15 & \textbf{26.7}  & \textbf{84.6}  & 13.3  & 57.1  \\
16 & \textbf{60.0}  & \textbf{99.8}  & 6.7   & 33.3  \\
17 & 0.0   & 0.0   & 0.0   & 0.0   \\
18 & 60.0  & 99.8  & \textbf{100.0} & \textbf{100.0} \\
19 & 40.0  & 95.8  & \textbf{60.0}  & \textbf{99.8}  \\
20 & 80.0  & \textbf{100.0} & \textbf{100.0} & \textbf{100.0} \\
21 & \textbf{26.7}  & \textbf{84.6}  & 20.0  & 73.6  \\
22-24 & 0.0   & 0.0   & 0.0   & 0.0   \\
\midrule
Avg & 62.2 & \textbf{77.7} & \textbf{66.1} & 75.9 \\
\# Solved & 19 & 19 & \textbf{20} & \textbf{20} \\
\bottomrule
\end{tabular}
}
\end{table}

Due to the model size and corresponding training data, these models could only successfully design no more than four analog circuits.
Some other models not shown in Table ~\ref{additional_results}, including Mistral-7B~\cite{jiang2023mistral} (2 solved), Llama 3-8B~\cite{llama3modelcard} (1 solved), Qwen-1.5-110B~\cite{bai2023qwen} (2 solved), Llama 2-7B, 13B~\cite{touvron2023llama} (0 solved).

\begin{figure}[!ht]
  \centering
  \includegraphics[width=\textwidth]{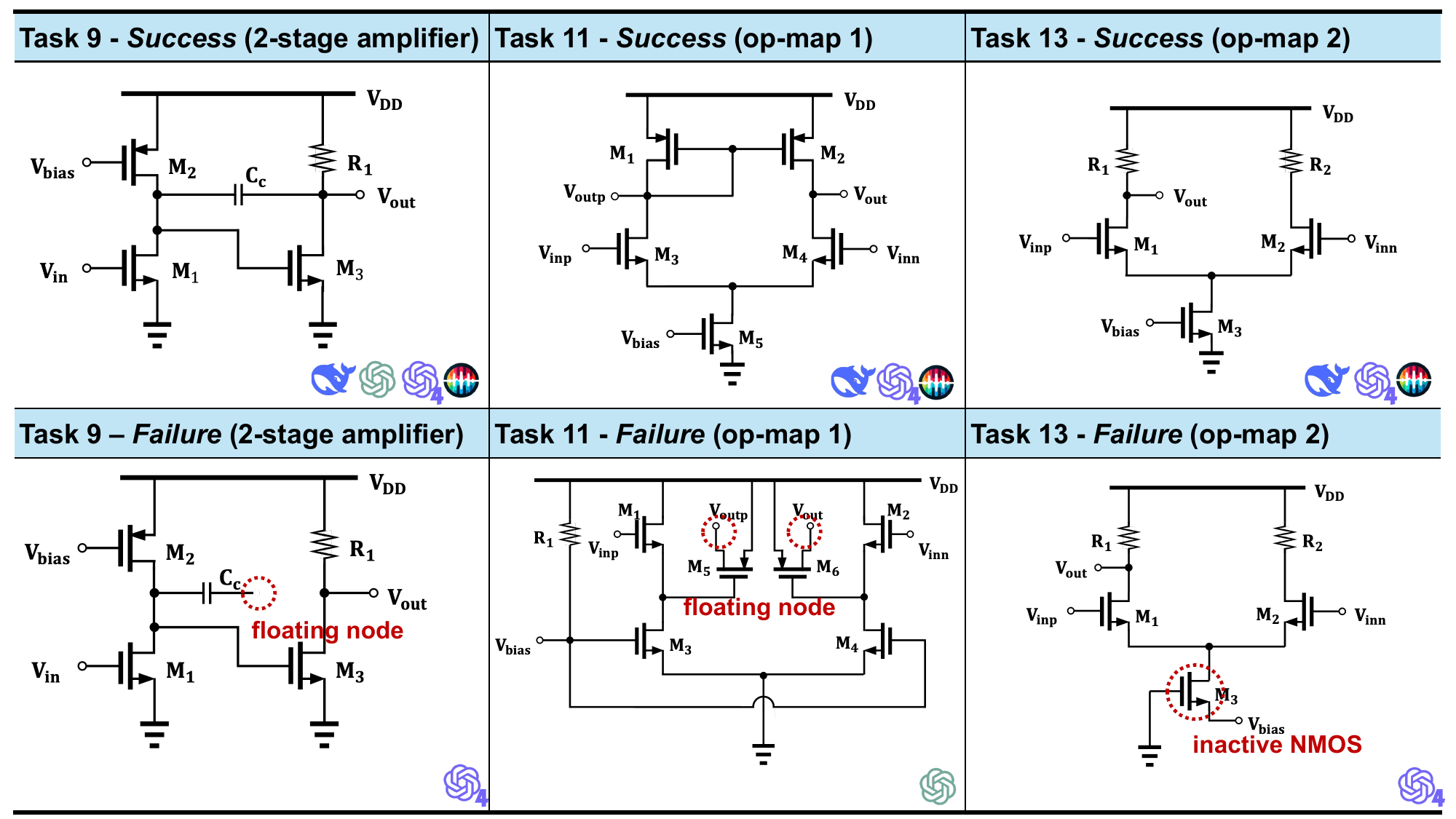}
  \caption{\small{\textbf{Additional visualization.} The LLM model used to design this circuit is detailed in the lower right corner.}
}
  \label{visualization_app}
\end{figure}

\paragraph{Additional Visualization.}
Additional visualizations are presented in Fig.~\ref{visualization_app}.

\begin{figure}[!ht]
  \centering
  \includegraphics[width=0.6\textwidth]{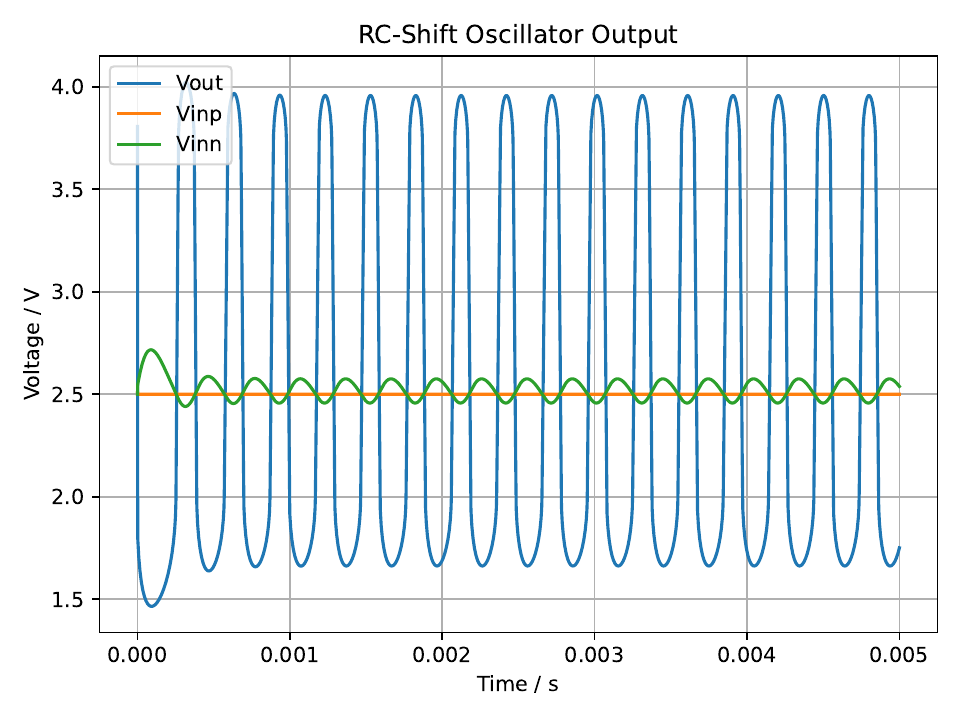}
  \caption{\small{\textbf{RC Shift Oscillator Simluation Result.} The simulation results of the RC shift oscillator, derived from a successful design by AnalogCoder, are shown in Fig.~\ref{visualization_app}.}
}
  \label{rc_shift}
\end{figure}

We also present a waveform generated during the simulation of an RC-shift oscillator implemented by the AnalogCoder, as shown in Fig.~\ref{rc_shift}.

\subsection{Error Bar Analysis}
\label{statistical_significance}

In our experiments, the primary evaluation metric utilized is \textit{Pass@k}. According to the formula, our calculations confirm that the \textit{Pass@k} is unbiased, as discussed in \cite{chen2021evaluating}.

\begin{figure}[!ht]
  \centering
  \includegraphics[width=0.5\textwidth]{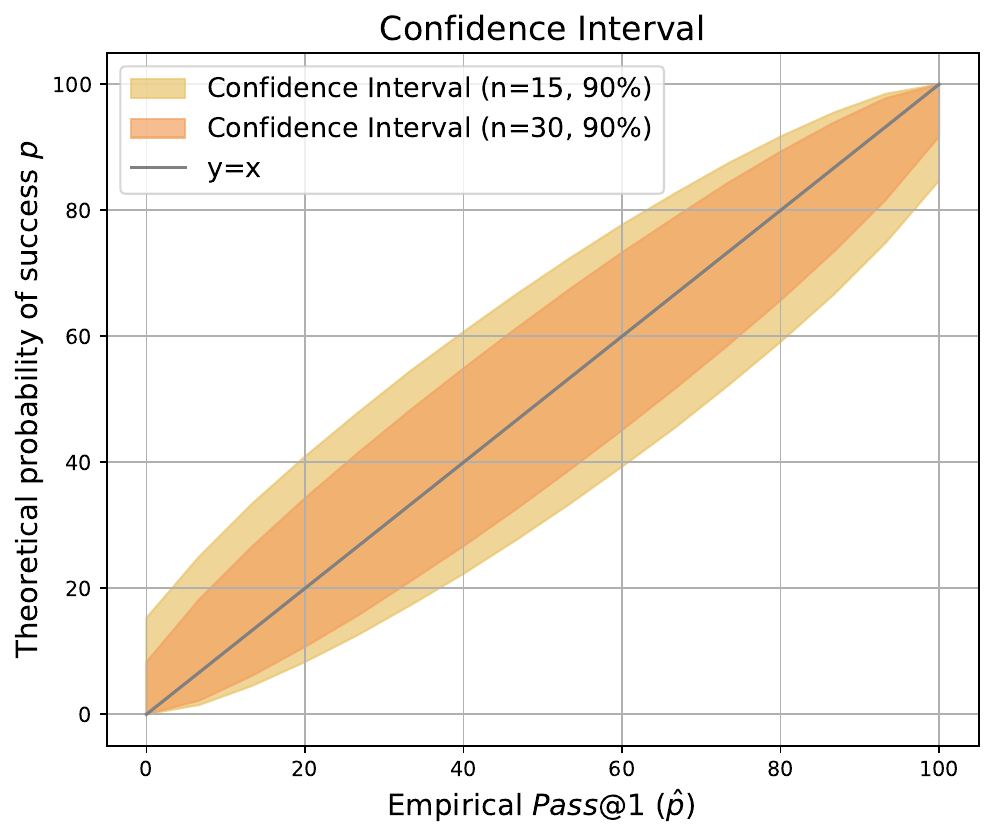}
  \caption{\small{\textbf{Confidence Interval Estimation.} }
}
  \label{confidence_interval}
\end{figure}

Moreover, we provide confidence intervals for estimating the theoretical value \( p \) using the experimental values of the design success rate (Pass@p), where \( n = 15, 30 \) represents the sample sizes used in our experiments, and the confidence level \( CI = 90\% \).
Since the experiments are conducted \( n \) times independently, the number of successful designs, \( c \), can be regarded as following a binomial distribution, donated as $c\sim \text{Bin}(n, p)$, where $P(c = k) = \binom{n}{k} p^k (1-p)^{n-k}$.
Therefore, we can employ the Wilson score interval~\cite{wilson1927probable} to estimate the confidence interval.
The results can be seen in Fig.~\ref{confidence_interval}.
The results indicate that increasing the number of sampling iterations can reduce the width of the confidence interval to a certain extent. Moreover, smaller values of \textit{Pass@1} tend to underestimate the theoretical value of \( p \); conversely, larger values of \textit{Pass@1} are likely to overestimate the theoretical \( p \). The number of trials in our experiment was determined by considering factors such as financial costs and computational resources.

\clearpage
\section{Related Work}
\paragraph{LLM for Code Generation.} 
Analogous to natural language, code generation can similarly be construed as a specific sequence generation task, one that can be effectively learned through language models. 
These models, trained on vast amounts of code data, have demonstrated remarkable capabilities in understanding and generating code across various programming languages \cite{chen2021evaluating,li2022competition,li2023starcoder,luo2023wizardcoder}.
However, one of the challenges that persist in this domain is the imbalance in the representation of different programming languages and domains within the training data. 
According to \citet{guo2024deepseek}, popular languages like Python, Java, JavaScript, and C account for around 60\% dataset.
This imbalance can lead to suboptimal performance when LLMs are tasked with generating code for underrepresented domains like robot control systems or circuit design.
To address this challenge, researchers have proposed the inclusion of domain-specific code samples during the training process. 
\citet{ahn2022can} incorporates robotics control codes into the training data.
\citet{liu2023chipnemo} domain-adaptive pre-trained the Llama2 model with Nvidia's internal design code data, including Verilog and VHDL~\cite{ashenden2010designer}.
Instead of training LLMs inflexibly, our work proposes a training-free, agent-based framework to enhance the circuit design capabilities of LLMs.

\paragraph{LLM for Electronic Design Automation.} Advancements in Electronic Design Automation (EDA) technologies have driven progress in the semiconductor industry~\cite{ICCAD20_DAMO,ICCAD21_DevelSet,DAC23_Nitho}, effectively assisting in chip design~\cite{zhong2023llm4eda, mirhoseini2021graph, lai2022maskplace, lai2023chipformer, lai2024scalable}.
The main applications facilitated by LLMs include assistance chatbots~\cite{liu2023chipnemo, han2023new}, HDL and script code  generation~\cite{blocklove2023chip, chang2023chipgpt, fu2023gpt4aigchip, thakur2023verigen, thakur2023autochip, liu2023verilogeval, lu2024rtllm, liu2023rtlcoder, pei2024betterv}, and code verification and analysis~\cite{yao2024hdldebugger, tsai2023rtlfixer}.
Assistant chatbots can help engineers with different hardware tasks by employing extensive hardware-related data for pre-training or fine-tuning. 
However, these models have not been made open source now due to the training involving proprietary data.
HDL and script code generation is currently one of the most extensively studied directions.
In the early stages of LLMs, due to the limited capabilities, most methods required interaction with human experts~\cite{blocklove2023chip, chang2023chipgpt}.
With the advent of more advanced LLMs such as GPT-4, researchers have begun to explore the fully automated generation of HDL or script code~\cite{lu2024rtllm, thakur2023verigen, fu2023gpt4aigchip}.
Some works have also enhanced the Verilog code generation capabilities by fine-tuning based on existing design data~\cite{liu2023rtlcoder} or by employing techniques such as Generative Discriminators~\cite{pei2024betterv}, and Monte Carlo Tree Search~\cite{delorenzo2024make}.
To provide a fair test dataset, researchers have introduced VerilogEval \cite{liu2023verilogeval} and RTLLM~\cite{lu2024rtllm}, open-source datasets aimed at assessing the capabilities in Verilog Code.
For code verification and analysis tasks, AssertLLM~\cite{fang2024assertllm} and SpecLLM~\cite{li2024specllm} use LLM for generating design specifications from natural languages.
HDLdebugger~\cite{yao2024hdldebugger} and RTLFixer~\cite{tsai2023rtlfixer} focus on the auto-fix error Verilog codes. 
However, these approaches remain applicable solely to digital circuits, while designing analog circuit components on chips continues to necessitate manual intervention.

\end{document}